\documentclass[runningheads]{llncs}
\usepackage{graphicx}
\usepackage{amsmath,amssymb} 
\usepackage[usenames, dvipsnames]{color}
\usepackage[width=122mm,left=12mm,paperwidth=146mm,height=193mm,top=12mm,paperheight=217mm]{geometry}

\usepackage{tabularx,ragged2e}
\usepackage{tabularx,ragged2e}
\usepackage{booktabs,calc}
\usepackage{multirow}
\usepackage{hhline}
\usepackage{array}
\usepackage{float}
\restylefloat{table}
\usepackage{makecell}
\usepackage{xcolor}
\usepackage{mathtools}
\usepackage{enumitem}
\usepackage{cite}

\usepackage[export]{adjustbox}
\mathchardef\mhyphen="2D
\usepackage{caption}
\usepackage[pagebackref=true,breaklinks=true,letterpaper=true,colorlinks,bookmarks=false]{hyperref}
\usepackage{siunitx}
\usepackage{subcaption}
\newcolumntype{Y}{>{\centering\arraybackslash}X}
\newcolumntype{s}{>{\hsize=.1\hsize}Y}
\newcolumntype{t}{>{\hsize=.2\hsize}Y}

\newcommand{\etal}{\textit{et al}.}
\newcommand{\ie}{\textit{i}.\textit{e}.}
\newcommand{\eg}{\textit{e}.\textit{g}.}

\begin{document}
\pagestyle{headings}
\mainmatter

\title{Hand Pose Estimation via Latent 2.5D Heatmap Regression} 
\titlerunning{Hand Pose Estimation via Latent 2.5D Heatmap Regression}
\authorrunning{Umar Iqbal \etal}
\author{Umar Iqbal$^{1,2}$, Pavlo Molchanov$^{1}$, Thomas Breuel$^{1}$ \\
Juergen Gall$^{2}$, Jan Kautz$^{1}$}
\institute{$^{1}$NVIDIA Research, $^{2}$University of Bonn, Germany}

\maketitle

\begin{abstract}
Estimating the 3D pose of a hand is an essential part of human-computer interaction. Estimating 3D pose using depth or multi-view sensors has become easier with recent advances in computer vision, however, regressing pose from a single RGB image is much less straightforward. The main difficulty arises from the fact that 3D pose requires some form of depth estimates, which are ambiguous given only an RGB image. In this paper we propose a new method for 3D hand pose estimation from a monocular image through a novel 2.5D pose representation. Our new representation estimates pose up to a scaling factor, which can be estimated additionally if a prior of the hand size is given. We implicitly learn depth maps and heatmap distributions with a novel CNN architecture. Our system achieves the state-of-the-art estimation of 2D and 3D hand pose on several challenging datasets in presence of severe occlusions. 

\end{abstract}
\section{Introduction}

Hand pose estimation from touch-less sensors enables advanced human machine interaction to increase comfort and safety. Estimating the pose accurately is a difficult task due to the large amounts of appearance variation, self occlusions and complexity of the articulated hand poses. 3D hand pose estimation escalates the difficulties even further since the depth of the hand keypoints also has to be estimated. To alleviate these challenges, many proposed solutions simplify the problem by using calibrated multi-view camera  systems~\cite{rehgECCV1994,campos2006cvpr,oikonomidis2010accv,rosales2001iccv,ballan_eccv2012,sridhar2014real,tzionas_ijcv2016,panteleris2017back,romero2017siggraphasia}, depth sensors~\cite{oikonomidis2011full,xu2013efficient,qian_cvpr2014,taylor2014cvpr,tang2014latent,tompson_tog2014,tang_iccv2015,makris2015cvpr,sridhar_cvpr2015,sun2015cvpr,oberweger_iccv2015,oberweger_cvpr2016,sridhar_eccv2016}, or color markers/gloves \cite{wang_TOG2009}. These approaches are, however, not 
very desirable due to their inapplicability in unconstrained environments. 
Therefore, in this work, we address the problem of 3D hand pose estimation from RGB images taken from the wild. 

Given an RGB image of the hand, our goal is to estimate the 3D coordinates of hand keypoints relative to the camera. 
Estimating the 3D pose from a monocular hand image is an ill-posed problem due to scale and depth ambiguities. Attempting to do so will either not work at all, or results in over-fitting to a very specific environment and subjects. We address these challenges by  decomposing the problem into two subproblems both of which can be solved without ambiguities. To this end, \textit{we propose a novel 2.5D pose representation and then provide a solution to reconstruct the 3D pose from 2.5D}. The proposed 2.5D representation is scale and translation invariant and can be easily estimated from RGB images. It consists of 2D coordinates of the hand keypoints in the input image, and scale normalized depth for each keypoint relative to the root (palm). We perform scale normalization of the depth values such that one of the bones always have a fixed length in 3D space. Such a constrained normalization allows us to directly reconstruct the scale normalized absolute 3D pose.

As a second contribution, \textit{we propose a novel CNN architecture to estimate the 2.5D pose from  images}. In the literature, there exists two main learning paradigms, namely heatmap regression~\cite{wei2016convolutional,newell2016eccv} and holistic pose regression~\cite{toshev2014deeppose,sun2017compositional}. 
Heatmap regression is now a standard approach for 2D pose estimation since it allows to accurately localize the keypoints in the image via per-pixel predictions. 
Creating volumetric heatmaps for 3D pose estimation~\cite{pavlakos2017volumetric}, however, results in very high computational overhead.
Therefore, holistic regression is a standard approach for 3D pose estimation, but it suffers from accurate 2D keypoint localization. 
Since the 2.5D pose representation requires the prediction of both the 2D pose and depth values, we propose a new heatmap representation that we refer to as \textit{2.5D heatmaps}. It consists of 2D heatmaps for 2D keypoint localization and a depth map for each keypoint for depth prediction. We design the proposed CNN architecture such that the 2.5D heatmaps do not have to be designed by hand, but are learned in a latent way. We do this by a \textit{softargmax} operation which converts the 2.5D heatmaps to 2.5D coordinates in a differentiable manner. 
The obtained 2.5D heatmaps are compact, invariant to scale and translation, and have the potential to localize keypoints with sub-pixel accuracy. 

We evaluate our approach on five challenging datasets with severe occlusions, hand object interactions and in-the-wild images. We demonstrate its effectiveness for both 2D and 3D hand pose estimation. The proposed approach outperforms state-of-the-art approaches by a large margin.
\vspace{-2mm}
\section{Related Work}
\vspace{-2mm}

Very few works in the literature have addressed the problem of 3D hand pose estimation from a single 2D image. The problem, however, shares several properties with human body pose estimation and many approaches proposed for human body can be easily adapted for hand pose estimation. Hence, in the following, we discuss the related works for 3D articulated pose estimation in general.  



\textbf{Model-based methods.} These methods represent the articulated 3D pose using 
a deformable 3D shape model. This is often formulated as an optimization problem, 
whose objective is to find the model's deformation parameters such that its projection 
is in correspondence with the observed image data~\cite{heap1996towards,wu2001iccv,Sigal_2010,gorce2011tpami,lu2003cvpr,bogo2016keep,Panteleris2018WACV}. 



\textbf{Search-based methods.} These methods follow a non-parametric approach and formulate 3D pose estimation as 
a nearest neighbor search problem from the large databases of 3D poses, where the matching is performed based on some 
low~\cite{athitsos2003cvpr,romero_icra2010} or high~\cite{Yasin_2016_CVPR,chen2017matching} level features extracted 
from the image.

\textbf{From 2D pose to 3D.} Earlier methods in this direction learn probabilistic 3D 
pose models from MoCap data and recover 3D pose by lifting the 2D keypoints
~\cite{Ramakrishna_2012,SimoSerraCVPR2013,Akhter:CVPR:2015,tome2017lifting}. 
More recent approaches, on the other hand, use deep neural 
networks to learn a mapping from 2D pose to 3D  \cite{Moreno_arxiv2016,martinez2017simple,Zimmermann2017ICCV}. 
Instead of 2D keypoint locations, \cite{Zimmermann2017ICCV,tekin2017fuse} use 2D heatmaps 
\cite{wei2016convolutional,newell2016eccv} as input and learn convolutional neural networks 
for 3D pose regression. The approach in~\cite{Zimmermann2017ICCV} is one of the first learning 
based methods to estimate 3D hand pose from a single RGB image. They use an existing 2D pose 
estimation model~\cite{wei2016convolutional} to first obtain the heatmaps of hand keypoints 
and feed them to another CNN that regresses a canonical pose representation and the camera 
view point.  

The aforementioned methods have the advantage that they do not necessarily require images 
with ground-truth 3D pose annotations for training,
their major drawback is that they 
cannot handle re-projection ambiguities (a joint with positive or negative depth will have the same 2D projections). Moreover, they are sensitive to errors in 2D 
image measurements and the required optimization methods are often prone to local minima due 
to incorrect initializations.

\textbf{3D pose from images.} These approaches aim to learn a direct mapping from RGB images 
to 3D pose~\cite{LiC14,zhou2016deep,GANeratedHands_2018}. While these methods can better handle 
2D projection ambiguities, their main downside is that they are prone to over-fitting 
to the views only present in training data. 
Thus, they require a large amount of training data with accurate 3D pose annotations. Collecting 
large amounts of training data in unconstrained environments is, however, infeasible. 
To this end, ~\cite{GANeratedHands_2018} proposes to use Generative 
Adversarial Networks~\cite{goodfellow2014generative} to convert synthetically generated hand images 
to look realistic. Other approaches formulate the problem in a multi-task setup to jointly 
estimate both 2D keypoint locations and 3D 
pose~\cite{popa2017CVPRmultitask,pavlakos2017volumetric,sun2017compositional,zhou2017weakly,nie2017iccv}. 
Our method also 
follows this paradigm. The closest work to ours are the approaches of 
\cite{sun2017compositional,zhou2017weakly,pavlakos2017volumetric} in that they also perform 2.5D 
coordinate regression. While the approach in~\cite{sun2017compositional} performs holistic pose 
regression with a fully connected output layer, \cite{zhou2017weakly} follows a hybrid approach and combines heatmap regression with holistic regression. Holistic regressions is shown to perform well for human 
body but fails in cases where very precise localization is required, \eg, finger-tips in case of hands. 
In order to deal with this, the approach in \cite{pavlakos2017volumetric} performs dense volumetric 
regression. This, however, substantially increases the model size, which in turn forces to work at a lower 
spatial resolution. Our approach, on the other hand, retains the input spatial resolution and allows one to 
localize hand keypoints with sub-pixel accuracy. It enjoys the differentiability and compactness of holistic 
regression-based methods, translation invariance of volumetric representations, while also providing 
high spatial output resolution. Moreover, in contrast to existing methods, it does not require hand-designed 
target heatmaps, which can arguably be sub-optimal for a particular problem, but rather implicitly learns a latent 2.5D heatmap representation and converts them to 2.5D coordinates in a differentiable way. 
 
Finally, note that given the 2.5D coordinates, the 3D pose has to be recovered. The existing approaches either make very strong assumptions such as the ground-truth location of the root~\cite{sun2017compositional} and the global scale of the hand in 3D is known~\cite{zhou2017weakly}, or resort to an approximate solution~\cite{pavlakos2017volumetric}. The approach~\cite{nie2017iccv} tries to directly regress the absolute depth from the cropped and scaled image regions which is a very ambiguous task. In contrast, our approach does not make any assumptions, nor does it try to solve any ambiguous task. Instead, we propose a scale and translation invariant 2.5D pose representation, which can be easily obtained using CNNs, and then provide an exact solution to obtain the absolute 3D pose up to a scaling factor and only approximate the global scale of the hand. 
\vspace{-1mm}

\begin{figure}[t]
\includegraphics[trim={0.0cm 0.cm 0.0cm 0},clip,scale=0.7]{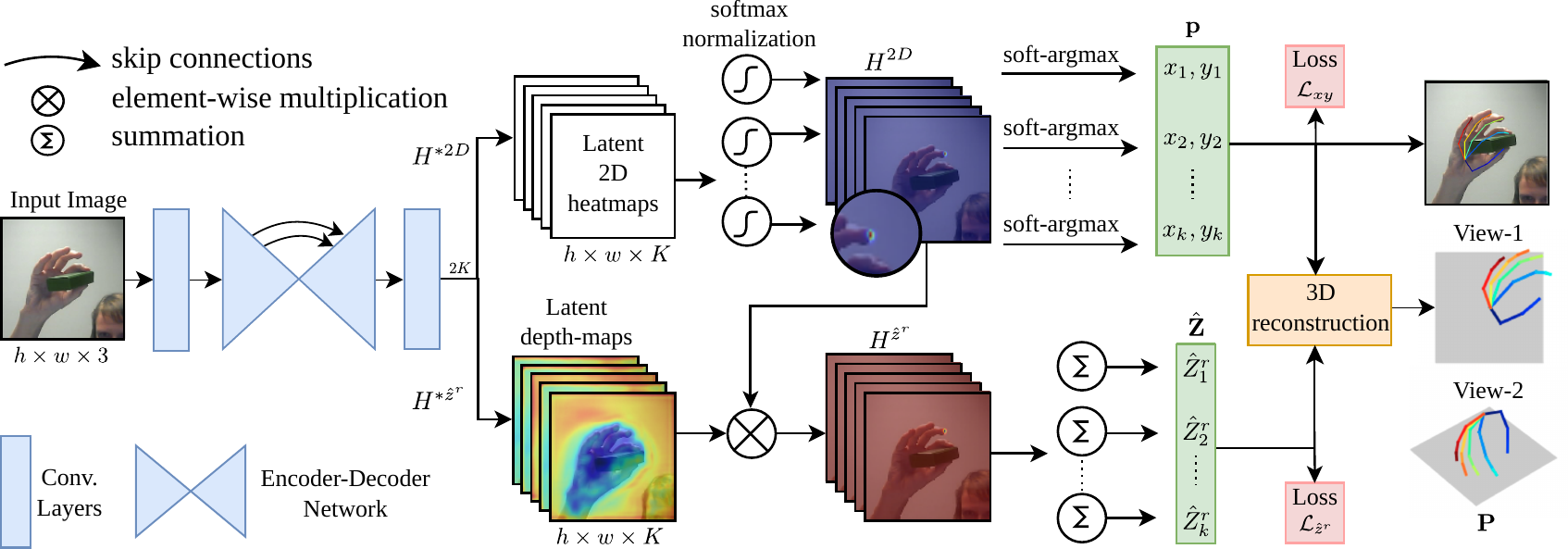}
\caption{Overview of the proposed approach. Given an image of a hand, the proposed CNN architecture produces latent 2.5D heatmaps containing the latent 2D heatmaps $H^{*2D}$ and latent depth maps $H^{*\hat{z}}$. The latent 2D heatmaps are converted to probability maps $H^{2D}$ using softmax normalization. The depth maps $H^{\hat{z}}$ are obtained by multiplying the latent depth maps $H^{*\hat{z}}$ with the 2D heatmaps. The 2D pose $\textbf{p}$ is obtained by applying spatial \textit{softargmax} on the 2D heatmaps, whereas the normalized depth values $\hat{\mathbf{Z}}^r$ are obtained by the summation of depth maps. The final 3D pose is then estimated by the proposed approach for reconstructing 3D pose from 2.5D.\vspace{-5mm}}
\label{fig:overview}
\end{figure}

\section{Hand Pose Estimation}
An overview of the proposed approach can be seen in Fig.~\ref{fig:overview}. Given an RGB image $\bf{I}$ of a hand, our goal is to estimate the 2D and 3D positions of all 
the $K=21$ keypoints of the hand. We define the 2D hand pose as $\mathbf{p}=\{p_k\}_{k \in K}$ 
and 3D pose as $\mathbf{P}=\{P_k\}_{k \in K}$, where $p_k=(x_k,y_k) \in \mathbb{R}^2$ 
represents the 2D pixel coordinates of the keypoint $k$ in image $\bf{I}$ and 
$P_k=(X_k,Y_k,Z_k) \in \mathbb{R}^3$ denotes the location of the keypoint in the 3D 
camera coordinate frame measured in millimeters. The Z-axis corresponds to 
the optical axis. Given the intrinsic camera parameters $\mathcal{K}$, the 
relationship between the 3D location $P_k$ and corresponding 2D projection $p_k$ can 
be written as follows under a perspective projection:
\begin{align}
    \label{eqt:projection3d2d}
    Z_k \begin{pmatrix}
          x_k \\           
          y_k \\
          1
    \end{pmatrix} = \mathcal{K} \begin{pmatrix}
          X_k \\           
          Y_k \\
          Z_k \\
          1
    \end{pmatrix} = \mathcal{K} \begin{pmatrix}
          X_k \\           
          Y_k \\
          Z_{root} + Z_k^r \\
          1
    \end{pmatrix}\quad k \in 1, \dots K
  \end{align}
where $k \in 1, \dots K$, $Z_{root}$ is the depth of the root keypoint, and 
$Z_k^r = Z_k - Z_{root}$ corresponds to the depth of the $k^{th}$ keypoint relative to the root. 
In this work we use palm of the hand as the root keypoint. 

\vspace{-2mm}
\subsection{2.5D Pose Representation}
\vspace{-1mm}
Given an image $\bf{I}$, we need to have a function $\mathcal{F}$, such 
that $\mathcal{F}: \bf{I} \to \mathbf{P}$, and the estimated 3D hand pose $\mathbf{P}$ can 
be projected to 2D with the camera parameters $\mathcal{K}$. However, predicting 
the absolute 3D hand pose in camera 
coordinates is infeasible due to irreversible geometry and scale 
ambiguities. We, therefore, choose a 2.5D pose representation which 
can be recovered from a 2D image without ambiguity, and provide a solution 
to recover the 3D pose from the 2.5D representation. We define the 2.5D 
pose as $\mathbf{P}^{2.5D}_k=\{P^{2.5D}_k\}_{k \in K}$, where 
$P^{2.5D}_k=(x_k, y_k,Z^r_k)$. The coordinates $x_k$ and $y_k$ are 
the image pixel coordinates of the $k^\mathrm{th}$ keypoint and $Z^r_k$ is 
its metric depth relative to the root keypoint. Moreover, in order to remove 
the scale ambiguities, we scale-normalize the 3D pose as follows:
\begin{equation}
\label{eqt:normalization}
\mathbf{\hat{P}} = \dfrac{C}{s} \cdot \mathbf{P},
\end{equation}
where $s=\Vert P_n - P_{parent(n)} \Vert_2$ is computed for each 3D pose 
independently. This results in a normalized 3D pose $\hat{\mathbf{P}}$ with a constant 
distance $C$ between a specific pair of keypoints $(n, parent(n))$. 
Subsequently, our normalized 2.5D representation for keypoint $k$ becomes 
$\hat{P}^{2.5D}_k=(x_k,y_k,\hat{Z}^r_k)$. 
Note that the 2D pose does not change due to the normalization, since the projection 
of the 3D pose remains the same. Such a normalized 2.5D representation has 
several advantages: it allows to effectively exploit image information;
it enables dense pixel-wise prediction (Sec.~\ref{sec:2.5D_pose_regression}); it allows us to perform multi-task learning so that multiple sources of training data can be used; and finally it
allows us to devise an approach to exactly recover the absolute 3D pose up to a 
scaling factor. We describe the proposed solution to obtain the
function $\mathcal{F}$ in Sec.~\ref{sec:2.5D_pose_regression}, while the 3D pose reconstruction from 2.5D pose 
is explained in the next section.

\subsection{3D Pose Reconstruction from 2.5D}\label{sec:3D_reconstruction}
\vspace{-1mm}
Given a 2.5D pose $\hat{\mathbf{P}}^{2.5D} = \mathcal{F}(\bf{I})$, we need to find 
the depth $\hat{Z}_{root}$ of the root keypoint to reconstruct the scale normalized 3D pose 
$\hat{\mathbf{P}}$ using Equation~\eqref{eqt:projection3d2d}. While there exists many 3D 
poses that can have the same 2D projection, given the 2.5D pose and intrinsic camera parameters, there 
exists a unique 3D pose that satisfies 
\begin{equation}
 (\hat{X}_n - \hat{X}_m)^2 + (\hat{Y}_n - \hat{Y}_m)^2 + (\hat{Z}_n - \hat{Z}_m)^2 = C^2,
\end{equation}
where $(n,\ m\! =\! parent(n))$ is the pair of keypoints used for normalization in Equation~(\ref{eqt:normalization}).
The equation above can be rewritten in terms of the 2D projections $(x_n,y_n)$ and $(x_m, y_m)$ as follows:
\begin{equation} 
 (x_n\hat{Z}_n - x_m\hat{Z}_m)^2 + (y_n\hat{Z}_n - y_m\hat{Z}_m)^2 + (\hat{Z}_n - \hat{Z}_m)^2 = C^2.
\end{equation}
Subsequently, replacing $\hat{Z}_n$  and $\hat{Z}_m$ with $(\hat{Z}_{root} + \hat{Z}_n^{r})$ and 
$(\hat{Z}_{root} + \hat{Z}_m^{r})$, respectively, yields:
\begin{multline}
(x_n (\hat{Z}_{root} + \hat{Z}_n^{r}) - x_m(\hat{Z}_{root} + \hat{Z}_m^{r}))^2 + (y_n(\hat{Z}_{root} + \hat{Z}_n^{r}) - y_m(\hat{Z}_{root} + \hat{Z}_m^{r}))^2 \\
+ ((\hat{Z}_{root} + \hat{Z}_n^{r}) - (\hat{Z}_{root} + \hat{Z}_m^{r}))^2 = C^2.
\end{multline}
Given the 2.5D coordinates of both keypoints $n$ and $m$, $Z_{root}$ is the only unknown in the equation above. 
Simplifying the equation further leads to a quadratic equation with the following coefficients
\begin{align}
a &= (x_n - x_m)^2 + (y_n - y_m)^2  \textsc{\nonumber }\\ 
b &= \hat{Z}_{n}^{r} (x_n^2 + y_n^2 - x_nx_m - y_ny_m) + \hat{Z}_{m}^{r} (x_m^2 + y_m^2 - x_nx_m - y_ny_m) \\
c &= (x_n\hat{Z}_n^{r} - x_m\hat{Z}_m^{r})^2 + (y_n\hat{Z}_n^{r} - y_m\hat{Z}_m^{r})^2 + (\hat{Z}_{n}^{r} - \hat{Z}_{m}^{r})^2 - C^2.\nonumber 
\label{eqt:coefficients}
\end{align}
This results in two values for the unknown variable $Z_{root}$, one in front of the 
camera and one behind the camera. 
We choose the solution in front of the camera
\begin{equation}
\hat{Z}_{root} = 0.5(-b+\sqrt{b^2-4ac})/a.
\label{eqt:quadractic_eqt}
\end{equation}
Given the value of $Z_{root}$, $\hat{\mathbf{P}}^{2.5D}$, and the intrinsic camera parameters $\mathcal{K}$, the scale normalized 3D pose can be reconstructed by back-projecting the 2D pose $\mathbf{p}$ using Eq.~\eqref{eqt:projection3d2d}. In this paper, we use $C=1$, and use the distance between the first joint (metacarpophalangeal - MCP) of the index finger and palm (root) to calculate the scaling factor $s$. We choose these keypoints since they are the most stable in terms of 2D pose estimation. 
 
\subsection{Scale Recovery}
Up to this point, we have obtained the 2D and scale normalized 3D pose $\hat{\mathbf{P}}$ of the hand.
In order to recover the absolute 3D pose $\mathbf{P}$, we need to know the global scale of the hand. 
In many scenarios this can be known a priori, however, in case it is not available, 
we estimate the scale $\hat{s}$ by 
\begin{equation}
\label{eqt:scale_recovery}
\hat{s} = \underset{s} {\mathrm{argmin}} \sum_{k,l \in \mathcal{E}} (s \cdot \Vert \hat{P}_k - \hat{P}_l \Vert - \mu_{kl})^2,
\end{equation}
where $\mu_{kl}$ is the mean length of the bone between keypoints $k$ and $l$ in the training data, and $\mathcal{E}$ defines 
the kinematic structure of the hand.

\section{2.5D Pose Regression}
\vspace{-1mm}
\label{sec:2.5D_pose_regression}
In order to regress the 2.5D pose $\mathbf{\hat{P}}^{2.5D}$ from an RGB image of the hand, 
we learn the function $\mathcal{F}$ using a CNN. In this section, we first describe an alternative 
formulation of the CNN  (Sec.~\ref{sec:direct_heatmap_regression}) and then describe our proposed solution for regressing latent 2.5D 
heatmaps in Sec.~\ref{sec:latent_heatmap_regression}. In all formulations, we train the CNNs using a loss function $\mathcal{L}$ which consists 
of two parts $\mathcal{L}_{xy}$ and $\mathcal{L}_{\hat{Z}^r}$, each responsible for the 
regression of 2D pose and root-relative depths for the hand keypoints, respectively. 
Formally, the loss can be written as follows:
\begin{equation}
\label{eqt:loss}
\mathcal{L}(\hat{\mathbf{P}}^{2.5D}) = \mathcal{L}_{xy}(\mathbf{p}, \mathbf{p}_{gt}) + \alpha \mathcal{L}_{\hat{Z}^r}(\hat{\bf{Z}}^r, \hat{\bf{Z}}^{r,gt}),
\end{equation}
where $\hat{\bf{Z}}^r=\{\hat{Z}_k^r\}_{r\in K}$ and $\hat{\bf{Z}}^{r,gt}=\{\hat{Z}_k^{r,gt}\}_{r\in K}$ and $gt$ refers to ground-truth annotations. 
This loss function has the advantage that it allows to utilize multiple 
sources of training, \ie, in-the-wild images with only 2D pose annotations and constrained or synthetic
images with accurate 3D pose annotations. While $\mathcal{L}_{xy}$ is valid for all training samples, 
$\mathcal{L}_{\hat{Z}^r}$ is enforced only when the 3D pose annotations are available, otherwise it is 
not considered. 

\subsection{Direct 2.5D Heatmap Regression}
\vspace{-1mm}
\label{sec:direct_heatmap_regression}
Heatmap regression is the de-facto approach for 2D pose estimation \cite{tompson2015cvpr,wei2016convolutional,newell2016eccv,simon2017hand}. In contrast to holistic regression, heatmaps have the advantage of providing higher output resolution, which helps in accurately localizing the keypoints. However, they are scarcely used for 3D pose estimation since a 3D volumetric heatmap representation~\cite{pavlakos2017volumetric}  results in a high computational and storage cost.  

We, thus, propose a novel and compact heatmap representation, which we refer to as \textit{2.5D heatmaps}.  It consists of 2D heatmaps $H^{2D}$ for keypoint localization and depth maps  $H^{\hat{z}^r}$ for depth predictions. While the 2D heatmap  $H^{2D}_k$  represents the likelihood of the $k^{th}$ keypoint at each pixel location, the depth map $H^{\hat{z}^r}_k$ provides the scale normalized and root-relative depth prediction for the corresponding pixels. This representation of depth maps is scale and translation invariant and remains consistent across similar hand poses, therefore, it is significantly easier to be learned using CNNs. The CNN provides a $2K$ channel output with $K$ channels for 2D localization heatmaps $H^{2D}$ and $K$ channels for depth maps $H^{\hat{z }^r}$. The target heatmap $H_k^{2D,gt}$ for the $k^\mathrm{th}$ keypoint is defined as 
\begin{equation}
H^{2D,gt}_{k}(p) = \exp\left(- \dfrac{\Vert p -  p^{gt}_k\Vert}{\sigma^2}\right), \quad p \in \Omega
\end{equation}
where $p^{gt}_k$ is the ground-truth location of the $k^\mathrm{th}$ keypoint, $\sigma$ controls the standard deviation of the heatmaps and $\Omega$ is the set of all pixel locations in image~$\bf{I}$. Since the ground-truth depth maps are not available, we define them by
\begin{equation}
H^{\hat{z}^r}_{k} = \hat{Z}_k^{r,gt} \cdot H^{2D,gt}_{k}
\end{equation}
where $\hat{Z}_k^{r,gt}$ is the ground-truth normalized  root-relative depth value of the $k^\mathrm{th}$ keypoint. During inference, the 2D keypoint position is obtained as the pixel with the maximum likelihood 
\begin{equation}
\label{eqt:heatmap_argmax}
p_k = {\underset{p} {\mathrm{argmax}}~H_k^{2D}(p)},
\end{equation}
and the corresponding depth value is obtained as, 
\begin{equation}
\hat{Z}^r_k = H^{\hat{z}^r}_k(p_k). 
\end{equation}

\vspace{-5mm}
\subsection{Latent 2.5D Heatmap Regression}
\label{sec:latent_heatmap_regression}
The 2.5D heatmap representation as described in the previous section is, arguably, not the most optimal representation. 
First, the ground-truth heatmaps are hand designed and are not ideal, \ie, $\sigma$ remains fixed for all keypoints and cannot be learned due to indifferentiability of Eq.~\eqref{eqt:heatmap_argmax}. Ideally, it should be adapted for each keypoint, \eg, heatmaps should be very peaky for finger-tips while relatively wide for the palm. Secondly, the Gaussian distribution is a natural choice for 2D keypoint localization, but is not very intuitive for depth prediction, \ie, the depth stays roughly the same throughout the palm but is modeled as Gaussians. Therefore, we alleviate these problems by proposing a latent representation of 2.5D heatmaps, \ie, the CNN learns the optimal representation by minimizing a loss function in a differentiable way.  

\begin{figure}[t]
\setlength{\tabcolsep}{0.25mm}
  \centering
\scalebox{1}{
  \begin{tabular}{l c c c c c c c c c c c}
    \rotatebox{90}{\centering \tiny{~~~Direct}} &
    \includegraphics[height=0.094\linewidth]{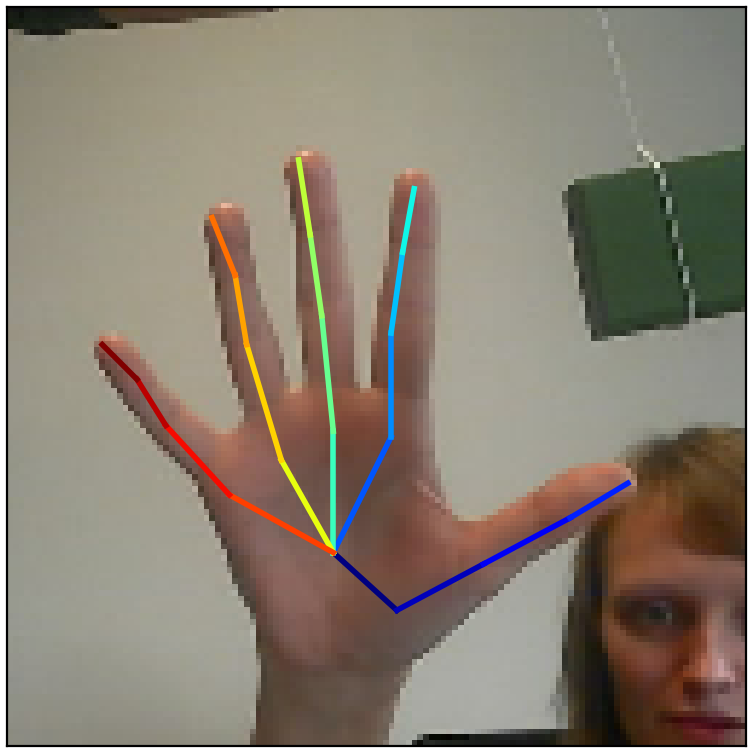} &
    \includegraphics[height=0.094\linewidth]{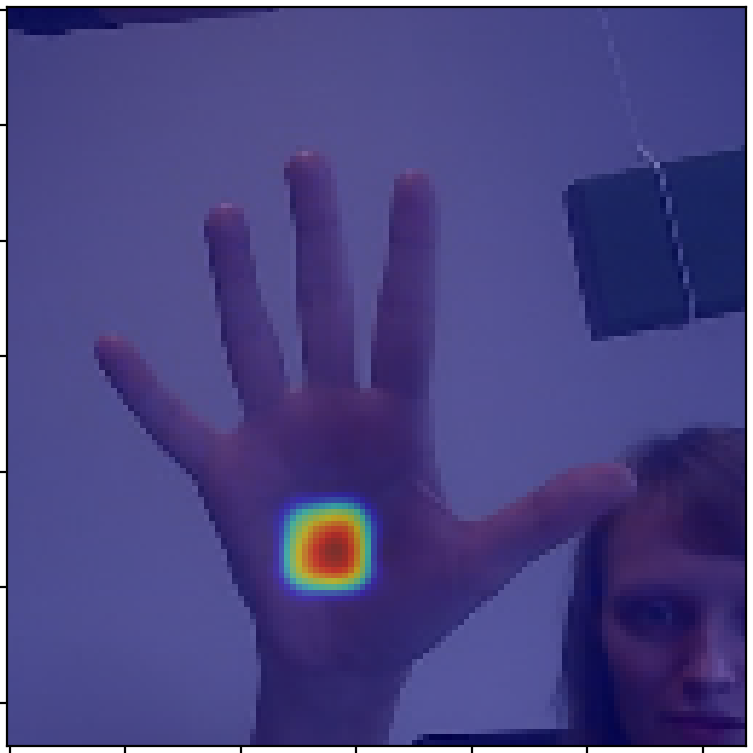} &
    \includegraphics[height=0.094\linewidth]{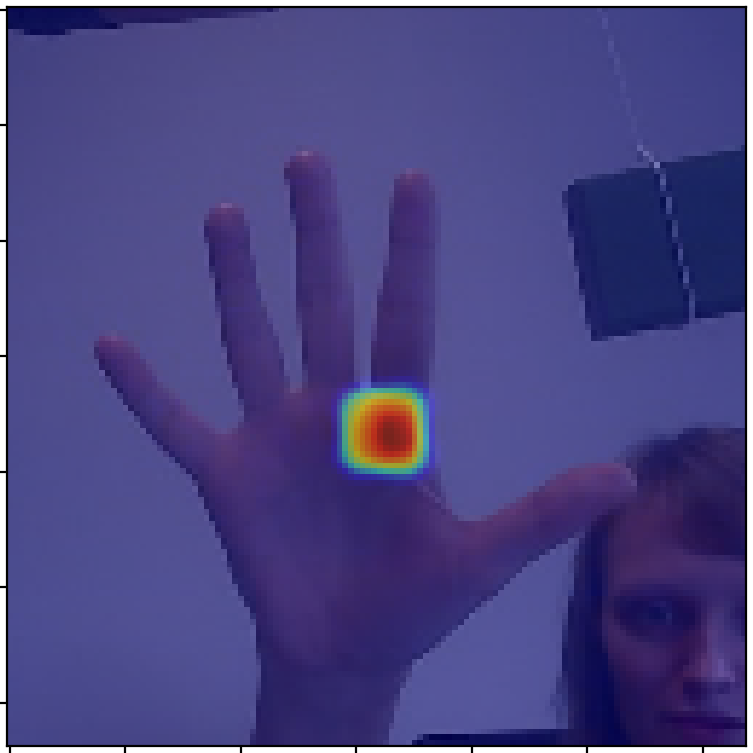} &
    \includegraphics[height=0.094\linewidth]{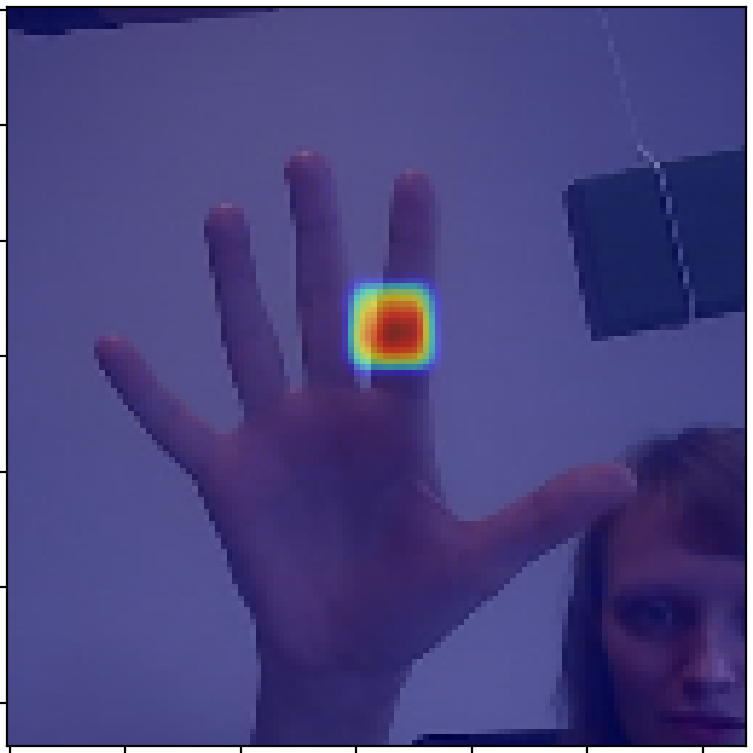} &
    \includegraphics[height=0.094\linewidth]{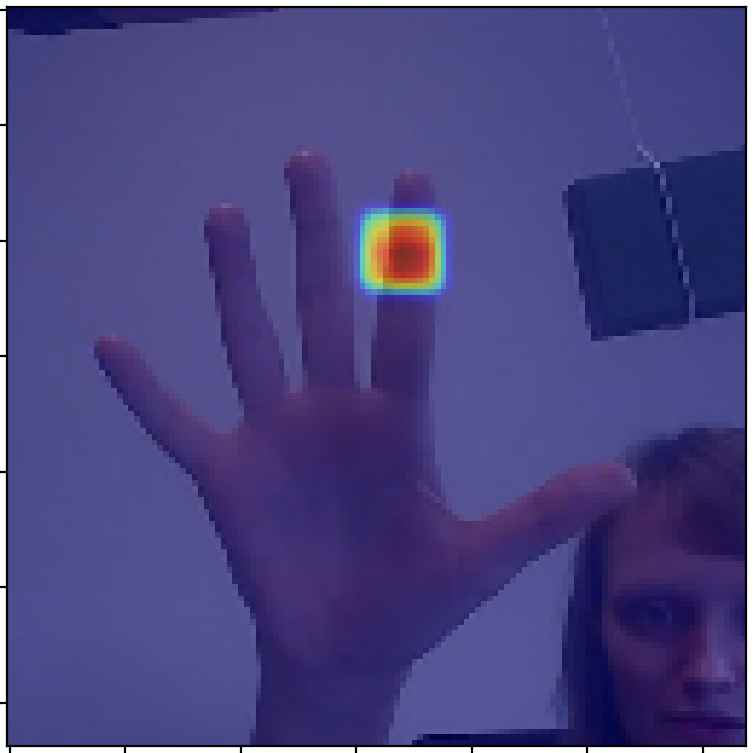} &
    \includegraphics[height=0.094\linewidth]{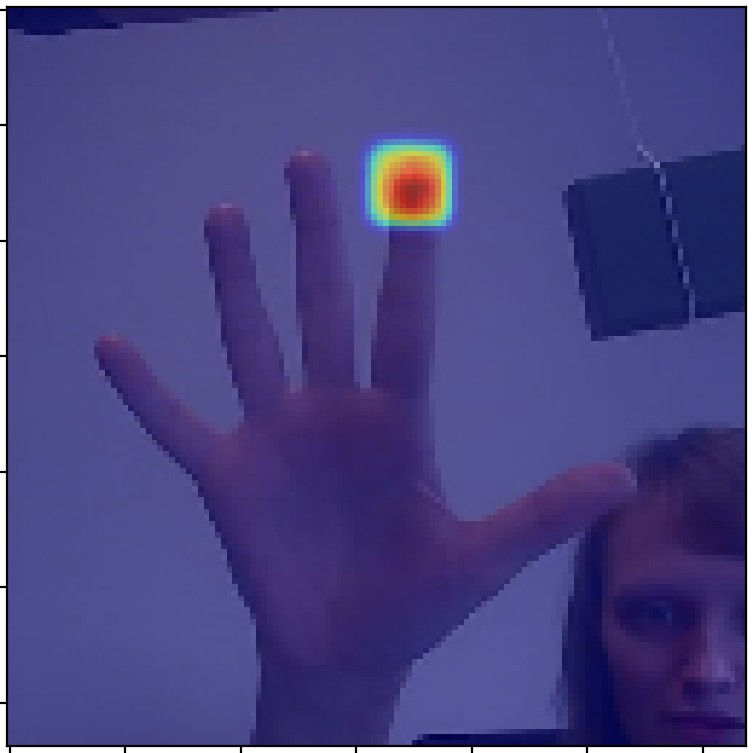}  & 
    \includegraphics[height=0.094\linewidth]{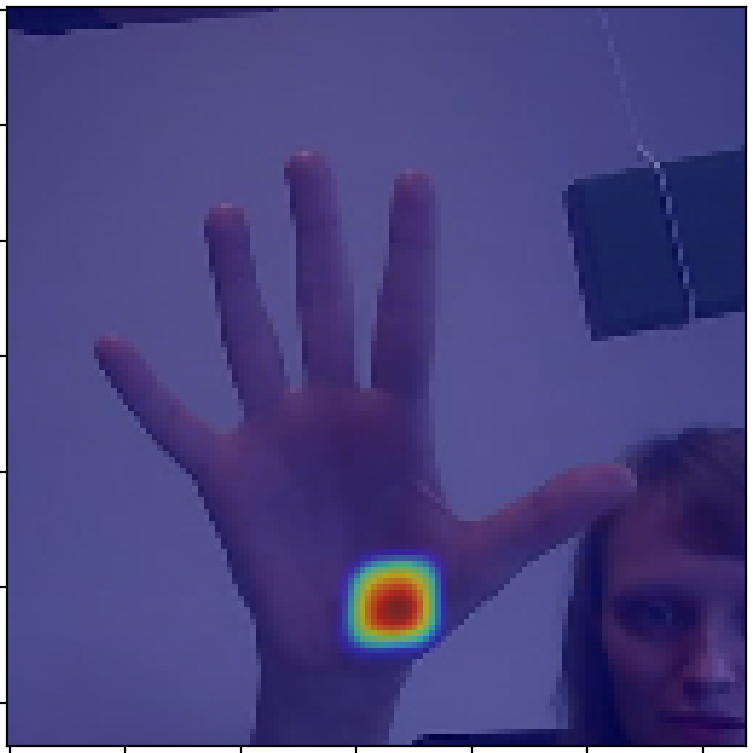} &
    \includegraphics[height=0.094\linewidth]{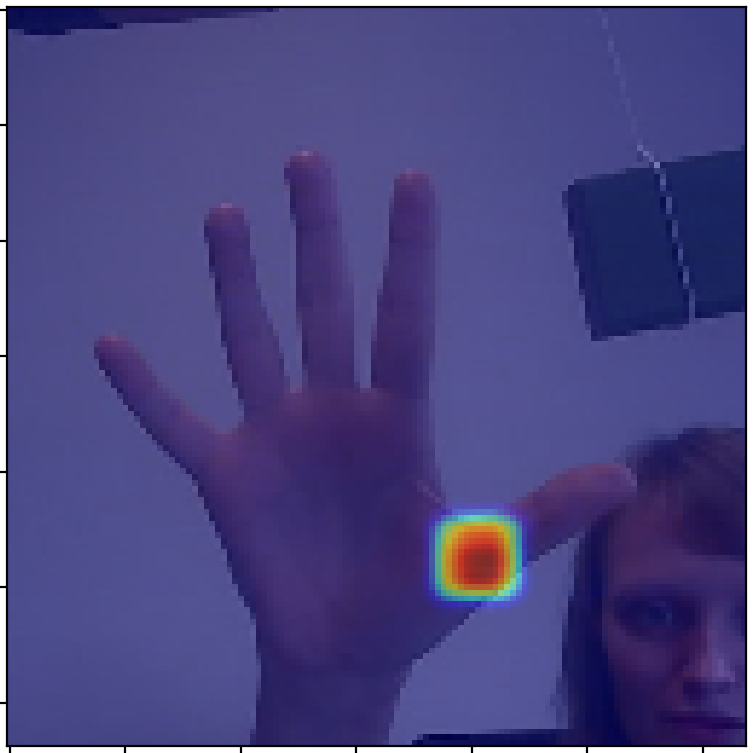} &
    \includegraphics[height=0.094\linewidth]{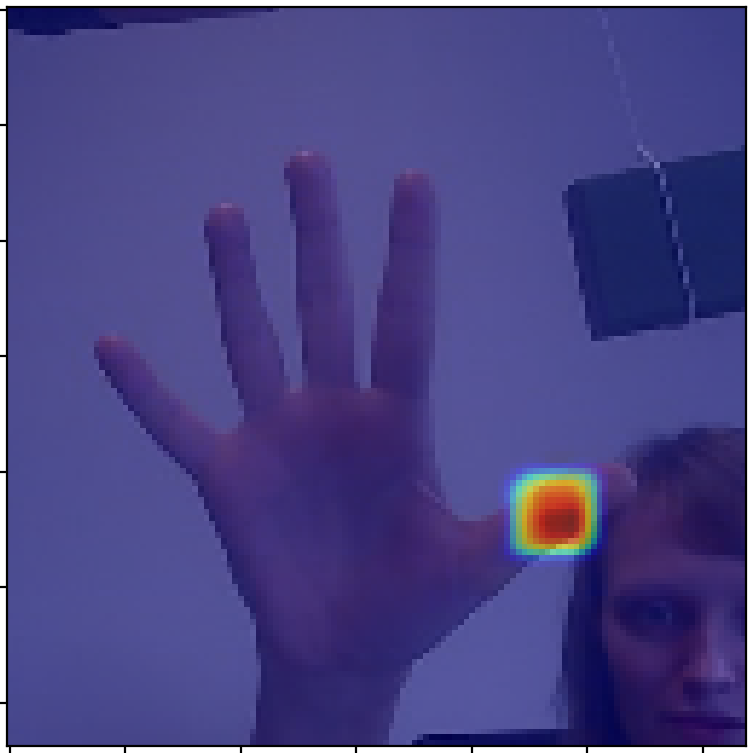} &
    \includegraphics[height=0.094\linewidth]{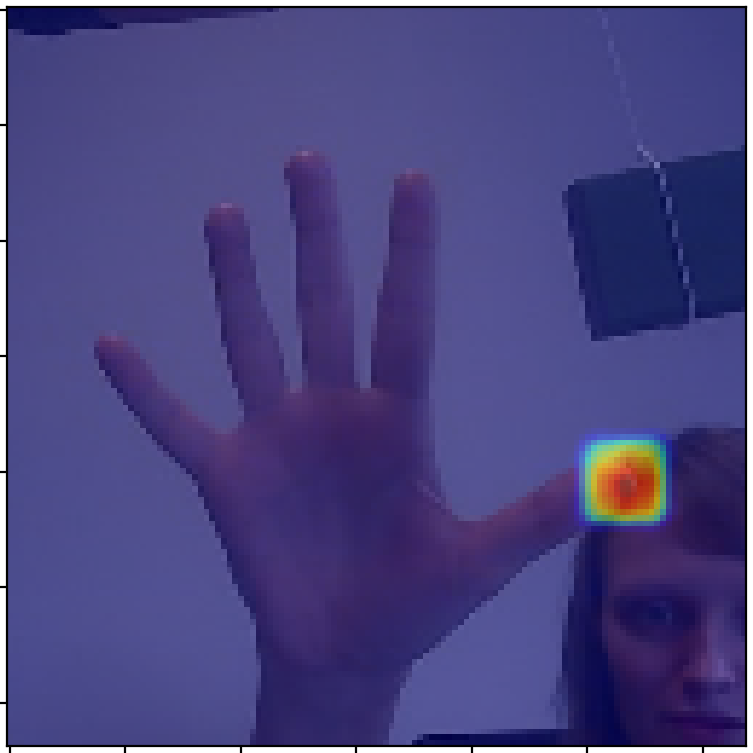}  &   
    \vspace{-.5mm}\\
    \rotatebox{90}{\centering \tiny{~~~Latent}} &
    \includegraphics[height=0.094\linewidth]{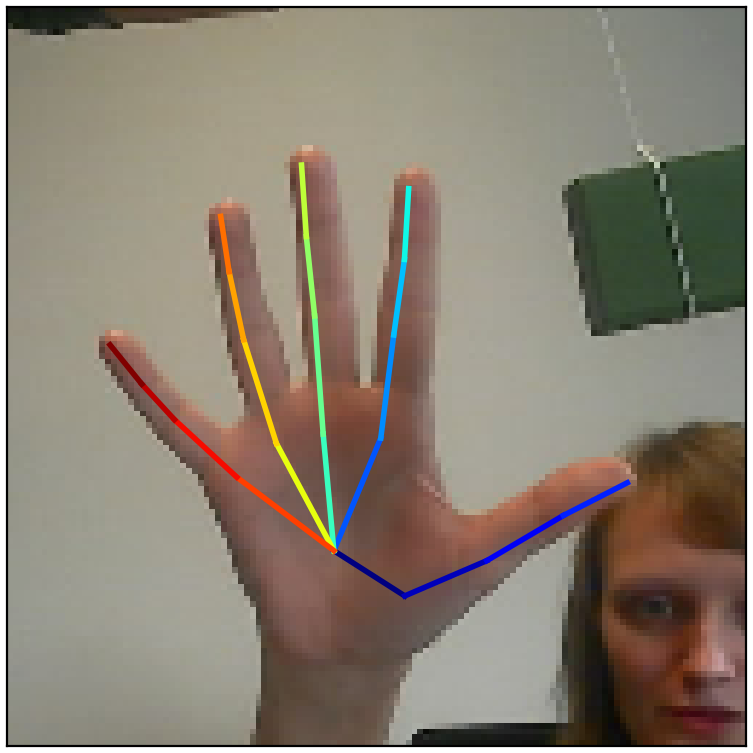} &
    \includegraphics[height=0.094\linewidth]{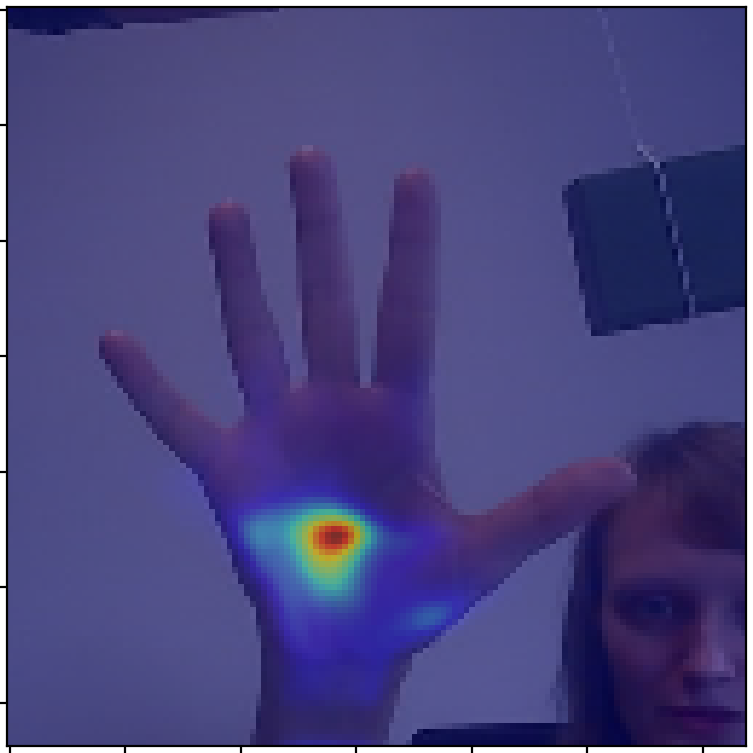} &
    \includegraphics[height=0.094\linewidth]{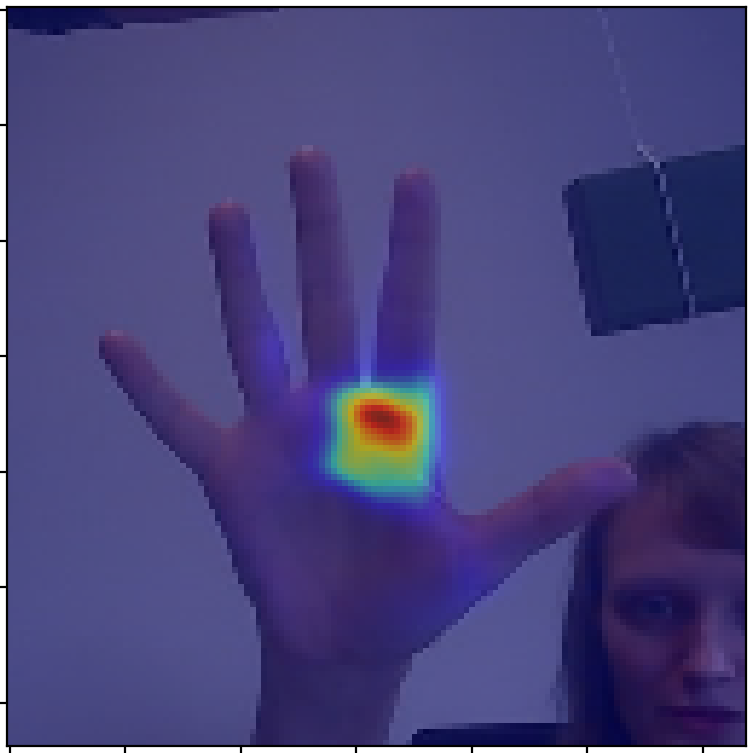} &
    \includegraphics[height=0.094\linewidth]{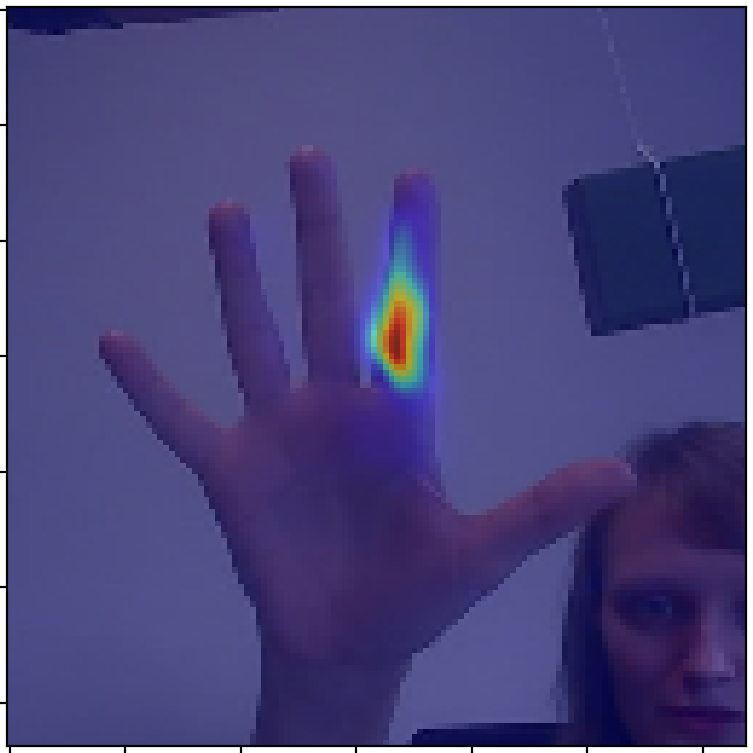} &
    \includegraphics[height=0.094\linewidth]{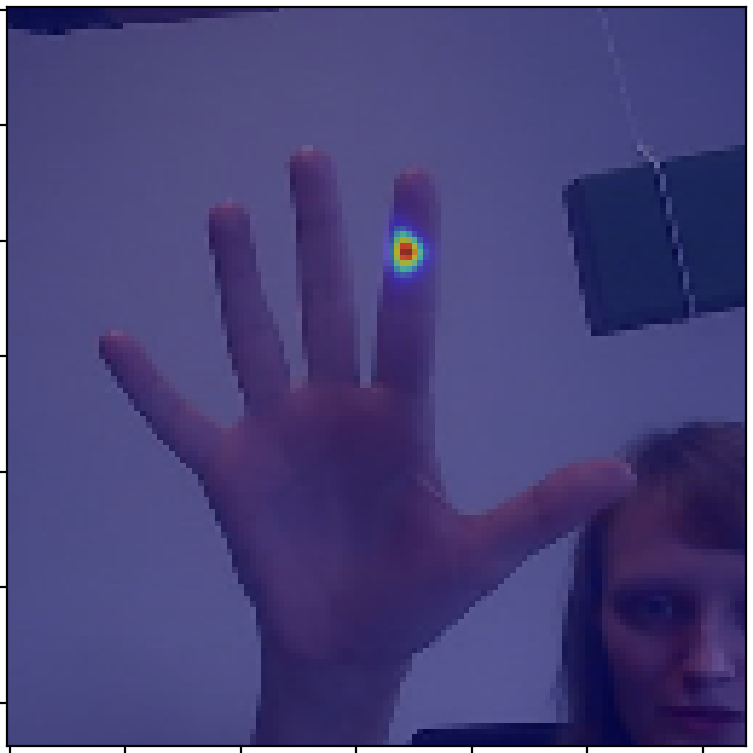} &
    \includegraphics[height=0.094\linewidth]{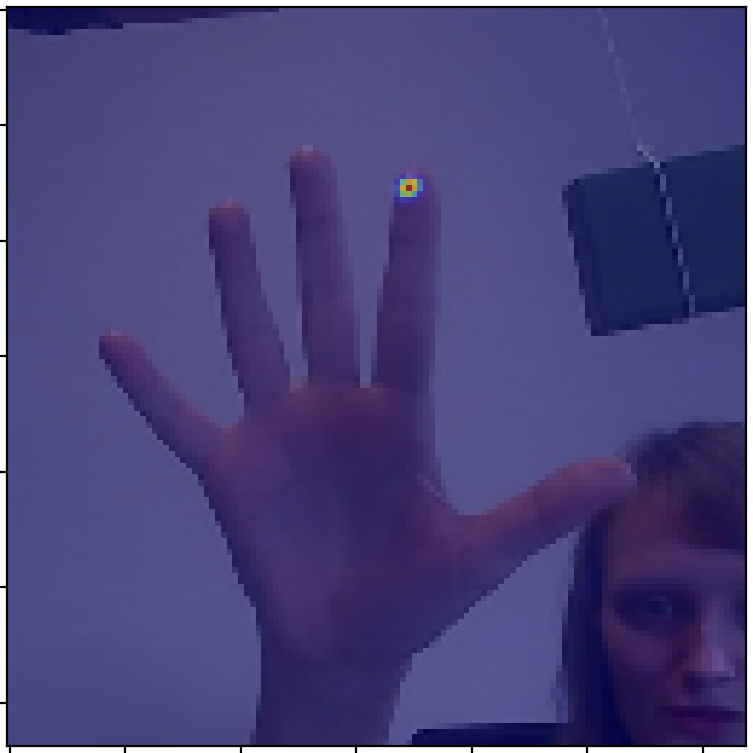}  & 
    \includegraphics[height=0.094\linewidth]{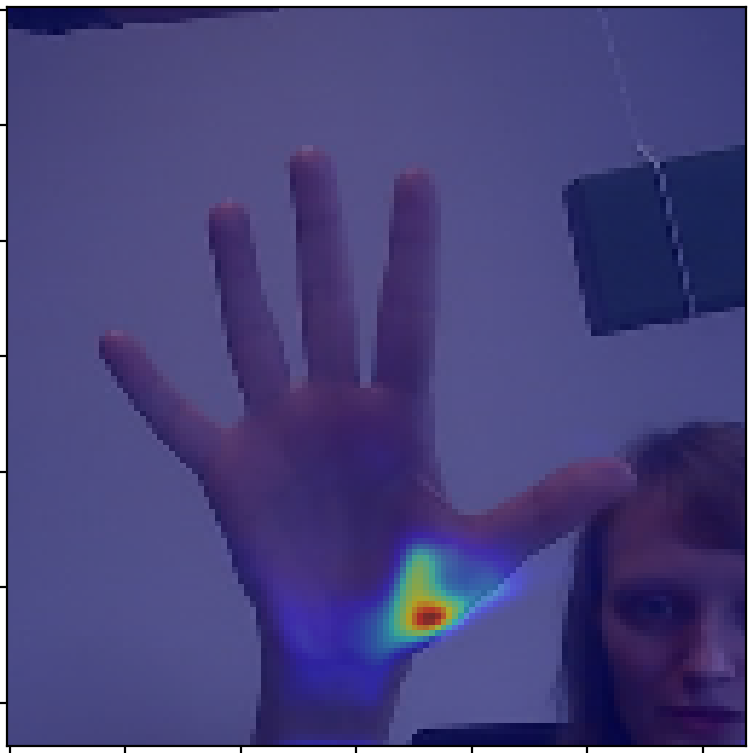} &
    \includegraphics[height=0.094\linewidth]{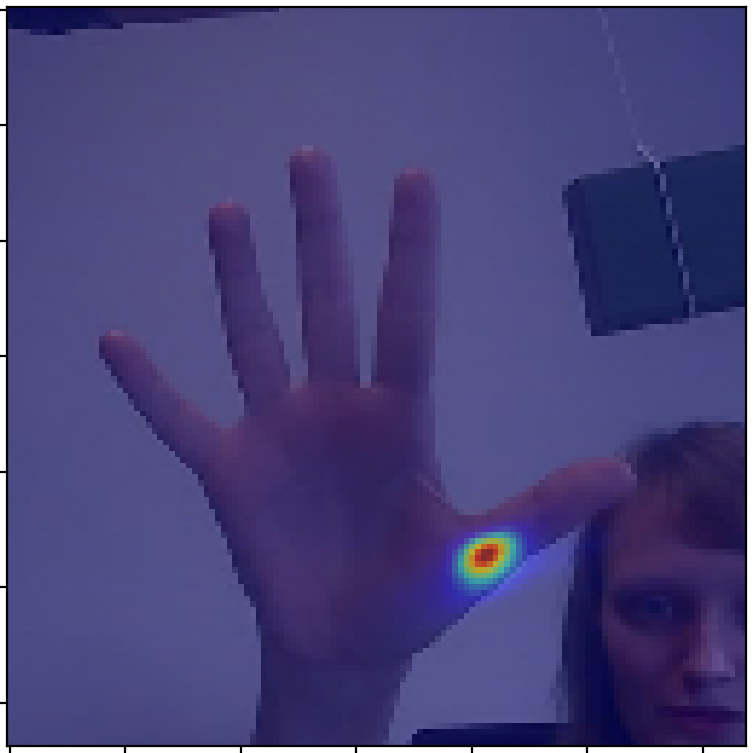} &
    \includegraphics[height=0.094\linewidth]{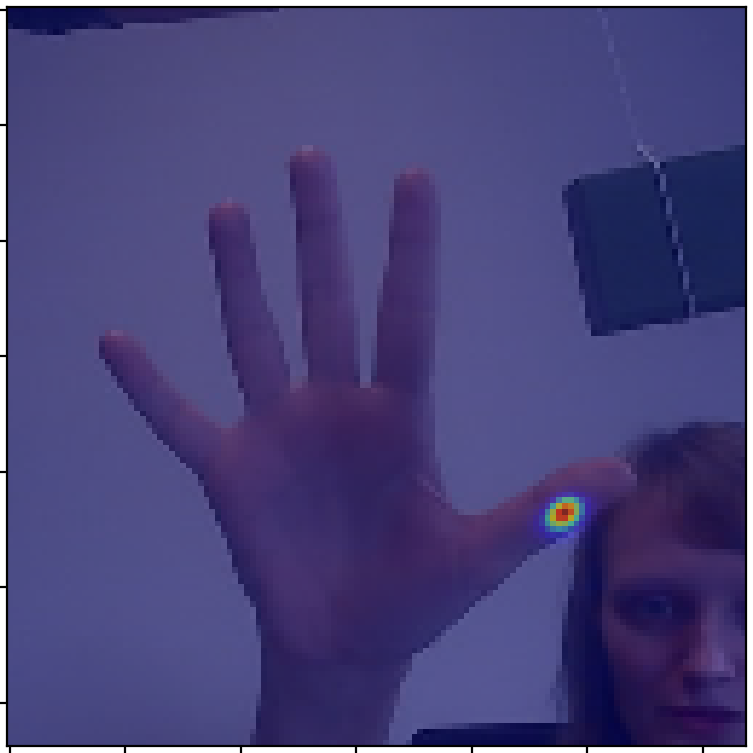} &
    \includegraphics[height=0.094\linewidth]{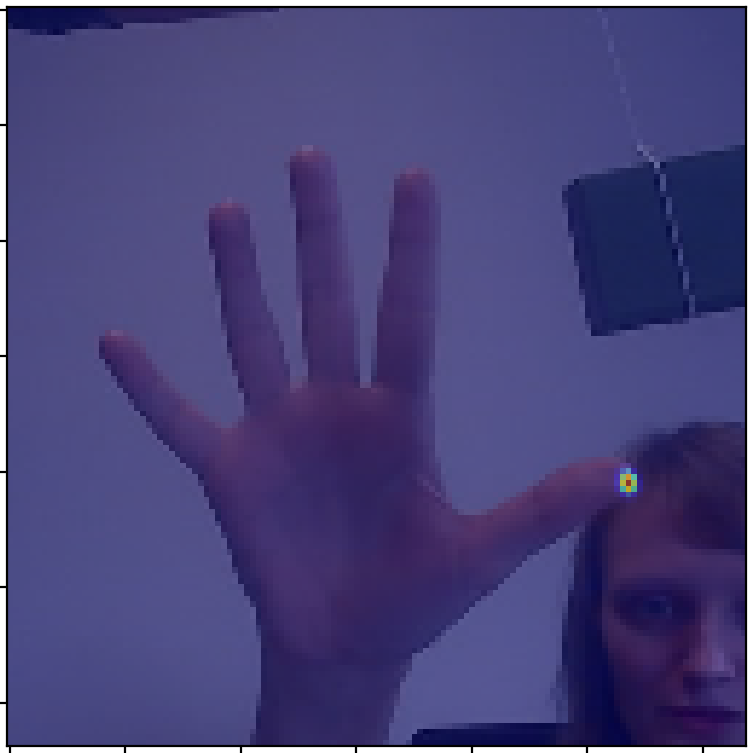}  &   
    \\
  \end{tabular}
}
\vspace{-3mm}
\caption{Comparison between the heatmaps obtained using direct heatmap regression (Sec.~\ref{sec:direct_heatmap_regression}) and the proposed latent heatmap regression approach (Sec.~\ref{sec:latent_heatmap_regression}). We can see how the proposed method automatically learns the spread separately for each keypoint, \ie, very peaky for finger-tips while a bit wider for the palm.\vspace{-5mm}}
\label{fig:qualitative_heatmaps_vs_latent_heatmap}
\end{figure}

To this end, we consider the $2K$ channel output of the CNN as latent variables $H_k^{*2D}$ and $H_k^{*\hat{z}^r}$ for 2D heatmaps and depth maps, respectively. We then apply spatial \textit{softmax} normalization to 2D heatmap $H_k^{*2D}$ of each keypoint $k$ to convert it to a probability map 
\begin{equation}
H^{2D}_k(p) = \dfrac{\exp(\beta_k  H^{*2D}_k(p))}{\sum_{p' \in \Omega} \exp(\beta_k H^{*2D}_k(p'))},
\end{equation}
where $\Omega$ is the set of all pixel locations in the input map $H_k^{*2D}$, and $\beta_k$ is the learnable parameter that controls the spread of the output heatmaps $H^{2D}$. Finally, the 2D keypoint position of the $k^\mathrm{th}$ keypoint is obtained as the weighted average of the 2D pixel coordinates as,
\begin{equation}
\label{eqt:softargmax}
p_k = \sum_{p \in \Omega} H^{2D}_k(p) \cdot p,
\end{equation}
while the corresponding depth value is obtained as the summation of the Hadamard product of $H^{2D}_k(p)$ and $H^{*\hat{z}^r}_k(p)$ as follows
\begin{equation}
\hat{Z}^r_k = \sum_{p \in \Omega} H^{2D}_k(p) \circ  H^{*\hat{z}^r}_k(p).
\end{equation}
A pictorial representation of this process can be seen in Fig.~\ref{fig:overview}. The operation in Eq.~\eqref{eqt:softargmax} is known as \textit{soft-argmax} in the literature \cite{chapelle2010}. Note that the computation of both the 2D keypoint location and the corresponding depth value is fully differentiable. Hence the network can be trained end-to-end, while generating latent 2.5D heatmap representation. In contrast to the heatmaps with fixed standard deviation in Sec. \ref{sec:direct_heatmap_regression}, the spread of the latent heatmaps can be adapted for each keypoint by learning the parameter $\beta_k$, while the depth maps are also learned implicitly without any ad-hoc design choices. A comparison between heatmaps obtained by direct heatmap regression and the ones implicitly learned by latent heatmap regression can be seen in Fig.~\ref{fig:qualitative_heatmaps_vs_latent_heatmap}. 

\vspace{-1mm}

\section{Experiments}
In this section, we evaluate the performance of the proposed approach in detail and also compare it with the state-of-the-art. For this, we use five challenging datasets -- namely, the Dexter+Object dataset ~\cite{sridhar_eccv2016}, the Ego-Dexter dataset~\cite{mueller_iccv2017}, the Stereo Hand Pose dataset~\cite{zhang2016stereo} dataset, the Rendered Hand Pose dataset~\cite{Zimmermann2017ICCV}, and the MPII+NZSL dataset \cite{simon2017hand}. The details of each dataset are as follows. 

\noindent\textbf{Dexter+Object (D+O).} The D+O dataset~\cite{sridhar_eccv2016} provides 6 test video sequences with $3145$ frames in total. All sequences are recorded using a static camera with a single person interacting with an object. The dataset provides both 2D and 3D pose annotations for the finger-tips of the left hand.\\
\noindent\textbf{EgoDexter (ED).} The ED dataset~\cite{mueller_iccv2017} provides both 2D and 3D pose annotations for 4  testing video sequences with $3190$ frames. The videos are recorded with body-mounted camera from egocentric viewpoints and contain cluttered backgrounds, fast camera motion, and complex interactions with various objects. Similar to D+O dataset, it only provides annotations for the finger-tips. In addition,~\cite{mueller_iccv2017} also provides the so called \textit{SynthHands} dataset containing synthetic images of hands from ego-centric views with accurate 3D pose annotations. The images are provided with chroma-keyed background, that we replace with random backgrounds from NYU depth dataset~\cite{Silberman:ECCV12} and use them as additional training data for testing on the ED dataset. \\
\noindent\textbf{Stereo Hand Pose (SHP).} The SHP dataset~\cite{zhang2016stereo} provides 2D and 3D pose annotations of $21$ keypoints for 6 pairs of stereo sequences with a total of 18000 stereo pairs of frames. The sequences record a single person performing a variety of gestures with different backgrounds and lighting conditions. \\
\noindent\textbf{Rendered Hand Pose (RHP).} The RHP dataset~\cite{Zimmermann2017ICCV} provides $41258$ and $2728$ images for training and testing, respectively. All images are generated synthetically using a blending software and come with accurate 2D and 3D annotations of $21$ keypoints. The dataset contains 20 different characters performing 39 actions with different lighting conditions, backgrounds, and camera viewpoints. \\
\noindent\textbf{MPII+NZSL.} The MPII+NZSL dataset~\cite{simon2017hand} provides $2800$ 2D hand pose annotations for in-the-wild images. The images are taken from YouTube videos of people performing daily life activities and New Zealand Sign Language exercises. The dataset is split into $2000$ and $800$ images for training and testing, respectively. In addition, ~\cite{simon2017hand} also provides additional training data that contains $14261$ synthetic images and $14817$ real images. The annotations for real images are generated automatically using multi-view bootstrapping. We refer to these images as MVBS in the rest of this paper. 
\vspace{-1mm}

\subsection{Evaluation Metrics}
For our evaluation on the D+O, ED, SHP, and RHP datesets, we use average \textit{End-Point-Error} (EPE) and the \textit{Area Under the Curve} (AUC) on the \textit{Percentage of Correct Keypoints} (PCK). We report the performance for both 2D and 3D hand pose where the performance metrics are computed in pixels and millimeters (mm), respectively. We use the publicly available implementation of evaluation metrics from~\cite{Zimmermann2017ICCV}. For the D+O and ED datasets, we follow the evaluation protocol proposed by~\cite{GANeratedHands_2018}, which requires estimating the absolute 3D pose with global scale. For SHP and RHP, we follow the protocol proposed by~\cite{Zimmermann2017ICCV}, where the root keypoints of the ground-truth and estimated poses are aligned before calculating the metrics. For the MPII+NZSL dataset, we follow~\cite{simon2017hand} and report \textit{head-normalized PCK} (PCKh) in our evaluation.
\vspace{-2mm}

\subsection{Implementation Details}
For 2.5D heatmap regression we use an Encoder-Decoder network architecture with skip connections~\cite{unet2015,newell2016eccv} and fixed number of channels (256) in each convolutional layer. The input to our model is a $128 \times 128$ image, which produces 2.5D heatmaps as output with the same resolution as the input image. Further details about the network architecture and training can be found in the appendix.
For all the video datasets, \ie, D+O, ED, SHP we use the YOLO detector~\cite{redmon2017yolo9000} to detect the hand in the first frame of the video, and  generate the bounding box in the subsequent frames using the estimated pose of the previous frame. We trained the hand detector using the training sets of all aforementioned datasets. 
\vspace{-2mm}

\subsection{Ablation Studies}

\begin{table}[t]
\centering
\scriptsize
\begin{tabularx}{\columnwidth}{Xccc|ccc}
\toprule
 \multirow{3}{6cm}{Method}  & \multicolumn{3}{c|}{2D Pose Estimation} &  \multicolumn{3}{c}{3D Pose Estimation} \\
 & \multirow{2}{1.2cm}{AUC $\uparrow$}    &  \multicolumn{2}{c|}{EPE (mm)}  &   \multirow{2}{1.2cm}{\centering AUC $\uparrow$}  &  \multicolumn{2}{c}{EPE (mm)}   \\
& & \multirow{1}{1.2cm}{median $\downarrow$ } & \multirow{1}{1.2cm}{mean $\downarrow$} & & \multirow{1}{1.2cm}{median  $\downarrow$} & \multirow{1}{1.2cm}{mean  $\downarrow$} \\
\midrule
\multicolumn{7}{c}{Comparison with baselines} \\
\midrule
Holistic 2.5D reg. 					& 0.41  & 17.34		& 22.21 	&  0.54 	&  42.76 	&  47.80 \\
Direct 2.5D heatmap reg. 			& 0.57 	& 10.33 	& 21.63 	&  0.55 	&  36.97 	&  52.33 \\
Latent 2.5D heatmap reg. (Ours)  	& 0.59 	& 9.91 		& 16.67 	&  0.57 	&  39.62 	&  45.54 \\
\midrule
\multicolumn{7}{c}{Impact of training data} \\
\midrule
\multicolumn{7}{l}{Latent 2.5D heatmap regression trained with} \\
~~~~SHP~\cite{zhang2016stereo}~+~RHP~\cite{Zimmermann2017ICCV} 	& 0.59 	& 9.91 		& 16.67 	&  0.57 	&  39.62 	&  45.54 \\
~~~~~~+~MPII~+~NZSL~\cite{simon2017hand} 						& 0.67 	& 9.07  	&  10.65 	&  0.68 	&  28.11 	&  32.78\\
~~~~~~~~+~MVBS~\cite{simon2017hand} 							& 0.68 	& 8.84 		&  10.45 	&  0.68 	&  27.27 	&  32.75\\
\midrule
\multicolumn{7}{c}{Performance after removing labeling discrepancy} \\
\midrule
Latent heatmap 2.5D reg.	& 0.76 	& 5.95 		&  7.97 	&  0.69 	&  26.56 	&  31.86\\
\bottomrule
\end{tabularx}
\vspace{-2mm}
\caption{Ablation studies. The arrows specify whether a higher or lower value for each metric is better. The first block compares the proposed approach of latent 2.5D heatmap regression with two baseline approaches. The second block shows the impact of different training data and the last block shows the impact due to differences in the annotations and using two stages. 
\vspace{-5mm}}
\label{tab:ablative_studies}
\end{table}

We evaluate the proposed method under different settings to better understand the impact of different design choices. We chose the D+O dataset for all ablation studies, mainly because it does not have any training data. Thus, it allows us to evaluate the generalizability of the proposed method. 
Finally, since the palm (root) joint is not annotated, it makes it compulsory to estimate the absolute 3D pose in contrast to the commonly used root-relative 3D pose. We use Eq.~\eqref{eqt:scale_recovery} to estimate the global scale of each 3D pose using the mean bone lengths from the SHP dataset. 

The ablative studies are summarized in Tab.~\ref{tab:ablative_studies}. We first examine the impact of different choices of CNN architectures for 2.5D pose regression. For holistic 2.5D pose regression, we use the commonly adopted~\cite{sun2017compositional} ResNet-50~\cite{he_cvpr2016} model. The details can be found in the appendix. We use the SHP and RHP datasets to train the models. Using a holistic regression approach results in an AUC of $0.41$ and $0.54$ for 2D and 3D  pose, respectively. Directly regressing the 2.5D heatmaps significantly improves the performance of 2D pose estimation ($0.41$ vs. $0.57$), while also raising the 3D pose estimation accuracy from $0.54$ AUC to $0.55$. Using latent heatmap regression improves the performance even further to $0.59$ AUC and $0.57$ AUC for 2D and 3D pose estimation, respectively. While the holistic regression approach achieves a competitive accuracy for 3D pose estimation, the accuracy for 2D pose estimation is inferior to the heatmap regression due to its limited spatial output resolution.

We also evaluate the impact of training the network in a multi-task setup. For this, we train the model with additional training data from~\cite{simon2017hand} which provides annotations for 2D keypoints only. First, we only use the $2000$ manually annotated real images from the training set of MPII+NZSL dataset. Using additional 2D pose annotations significantly improves the performance. Adding additional $15,000$ annotations of real images, automatically generated by multi-view bootstrapping~\cite{simon2017hand}, improves the performance only slightly. Hence, only $2000$ real images are sufficient to generalize the model trained on synthetic data to a realistic scenario. 

The annotations of the finger tips in the D+O dataset are slightly different than the other datasets. In the D+O dataset, the finger tips are annotated at the middle of the tips whereas other datasets annotate it at the edge of the nails. To remove this discrepancy, we shorten the last bone of the finger tip by $0.9$. Fixing the annotation differences results in further improvements, revealing the true performance of the proposed approach.

Finally, we also evaluate the impact of using multiple stages in the network, where each stage produces latent 2.5D heatmaps as output. While the first stage only uses the features extracted from the input image using the initial block of convolutional layers, each subsequent stage also utilizes the output of the preceding stage as input. This provides additional contextual information to the subsequent stages and helps in incrementally refining the predictions. Similar to~\cite{newell2016eccv,wei2016convolutional} we provide local supervision to the network by enforcing the loss at the output of each stage (see appendix for more details). Adding one extra stage to the network increases the 3D pose estimation accuracy from AUC $0.69$ to $0.71$, but decreases the 2D pose estimation accuracy from AUC $0.76$ to $0.74$. The decrease in 2D pose estimation accuracy is most likely due to over-fitting to the training datasets. Remember that we do not use any training data from the D+O dataset. In the rest of this paper, we always use networks with two stages unless stated otherwise. 

\vspace{-2mm}
\subsection{Comparison to State-of-the-Art}
\vspace{-1mm}

\begin{figure*}[t] 
   \centering
     \begin{subfigure}[t]{0.32\textwidth}
   	  \center
      \textbf{\tiny{Dexter+Object}}
	  \includegraphics[width=\columnwidth]{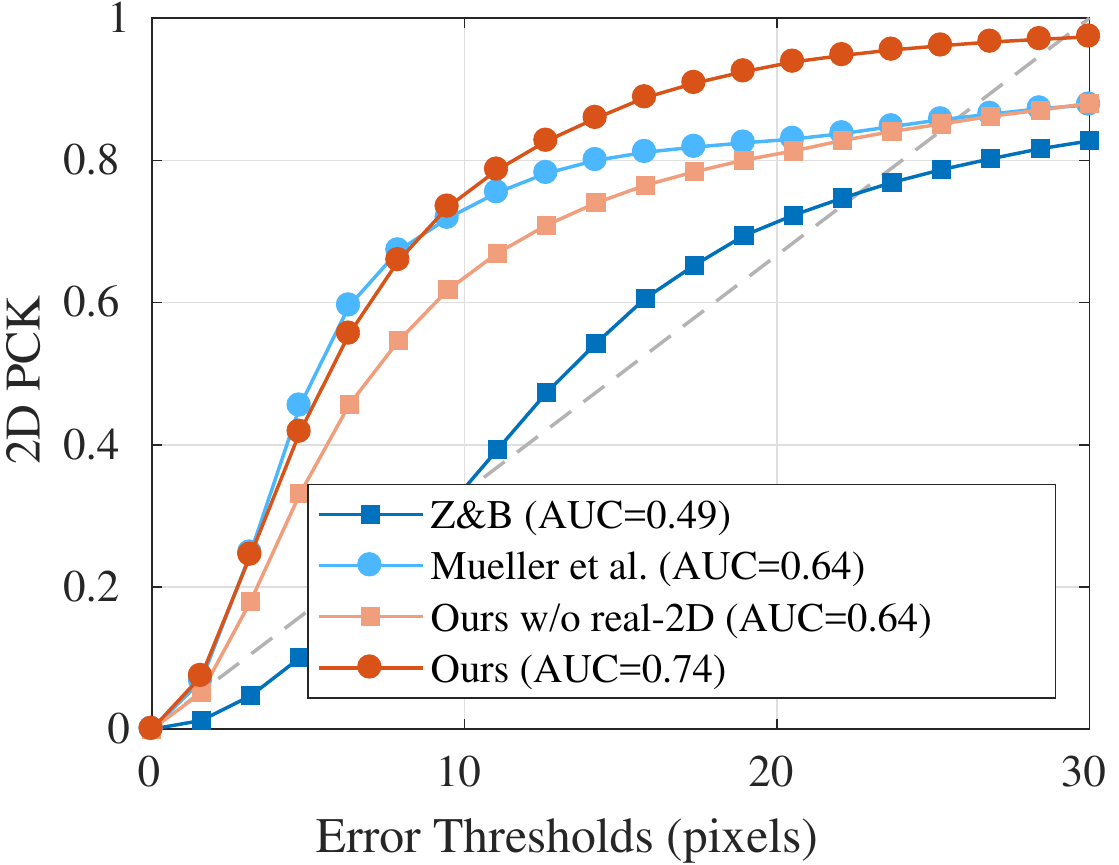}
	  \caption{2D PCK on D+O dataset.}
	  \label{fig:dexter_object_2d}
	\end{subfigure}
   \begin{subfigure}[t]{0.32\textwidth}
   	  \center
      \textbf{\tiny{Dexter+Object}}
	  \includegraphics[width=\columnwidth]{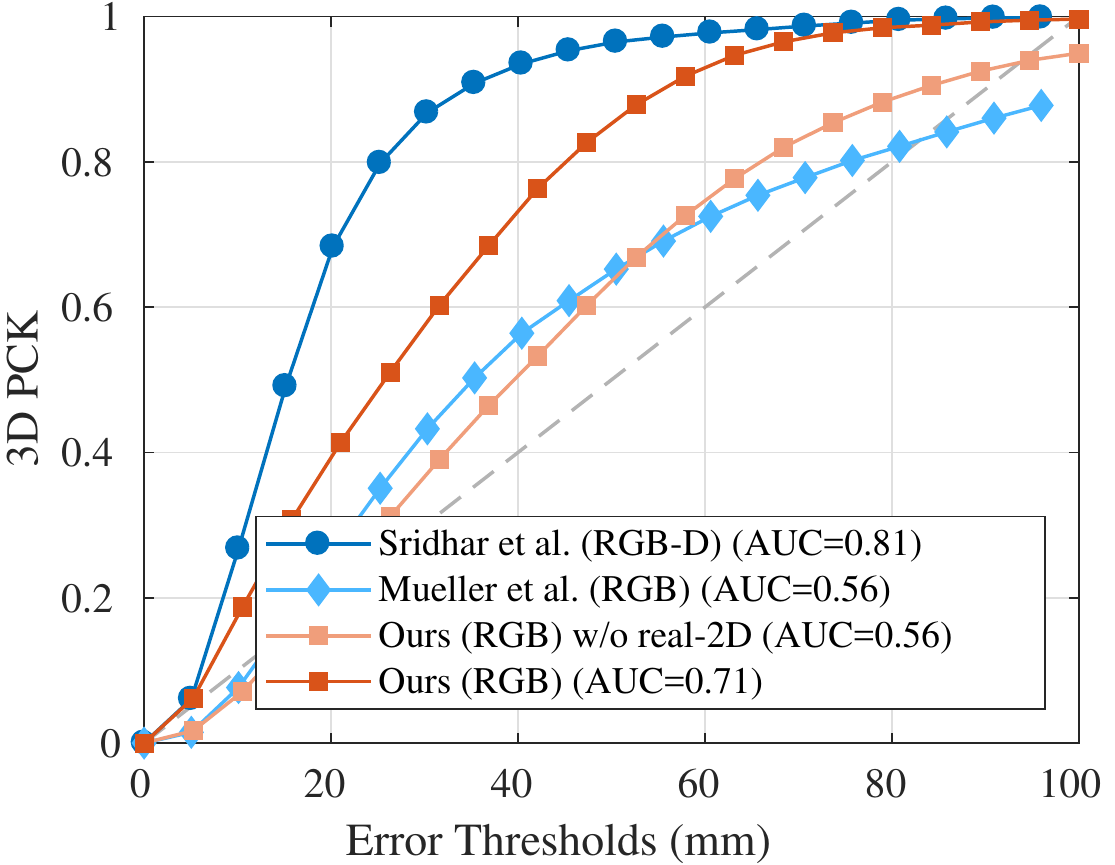}
	  \caption{3D PCK on D+O dataset.} 
	  \label{fig:dexter_object_3d}
	\end{subfigure}
    \begin{subfigure}[t]{0.31\textwidth}
     \center
     \textbf{\tiny{Stereo Hand Pose}} 
    \includegraphics[width=\columnwidth]{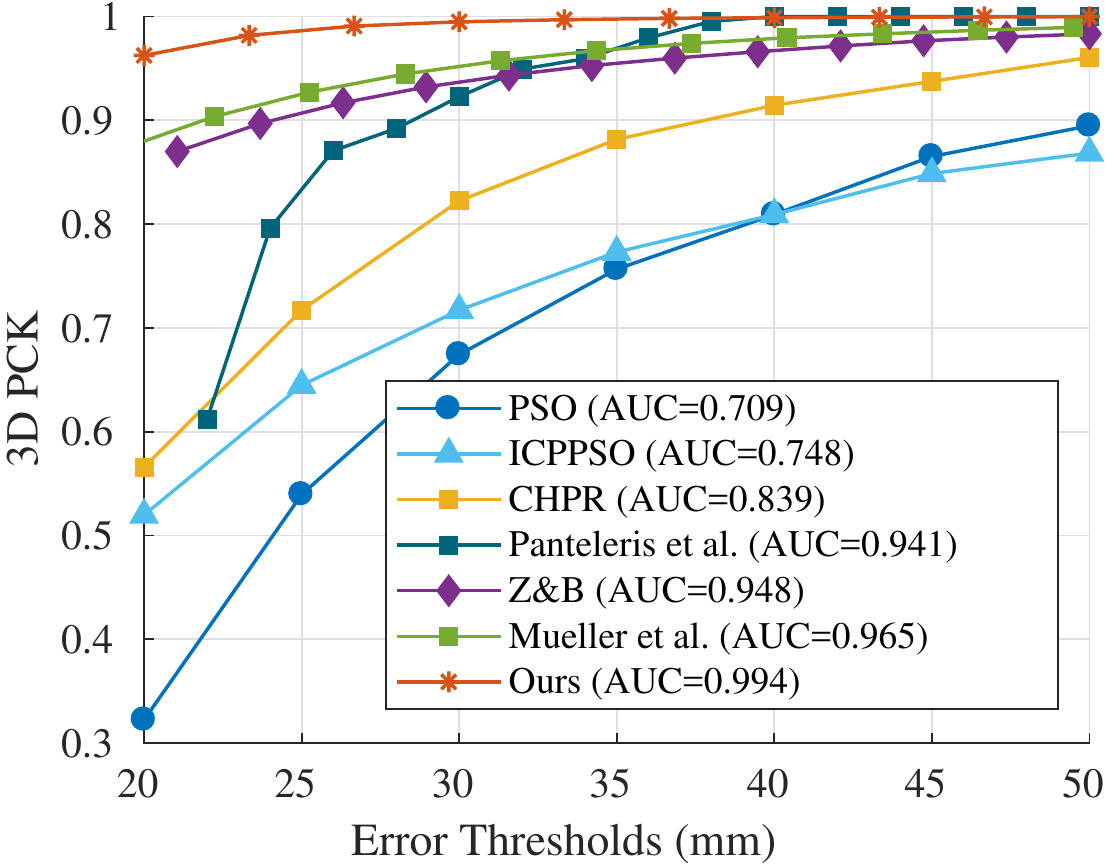}
     \caption{3D PCK on SHP dataset. } 
  	\label{fig:stereodataset}
	\end{subfigure}
     \centering
     \begin{subfigure}[t]{0.31\textwidth}
      \tiny
   	  \center
       \textbf{\tiny{EgoDexter}} 
	  \includegraphics[width=\columnwidth]{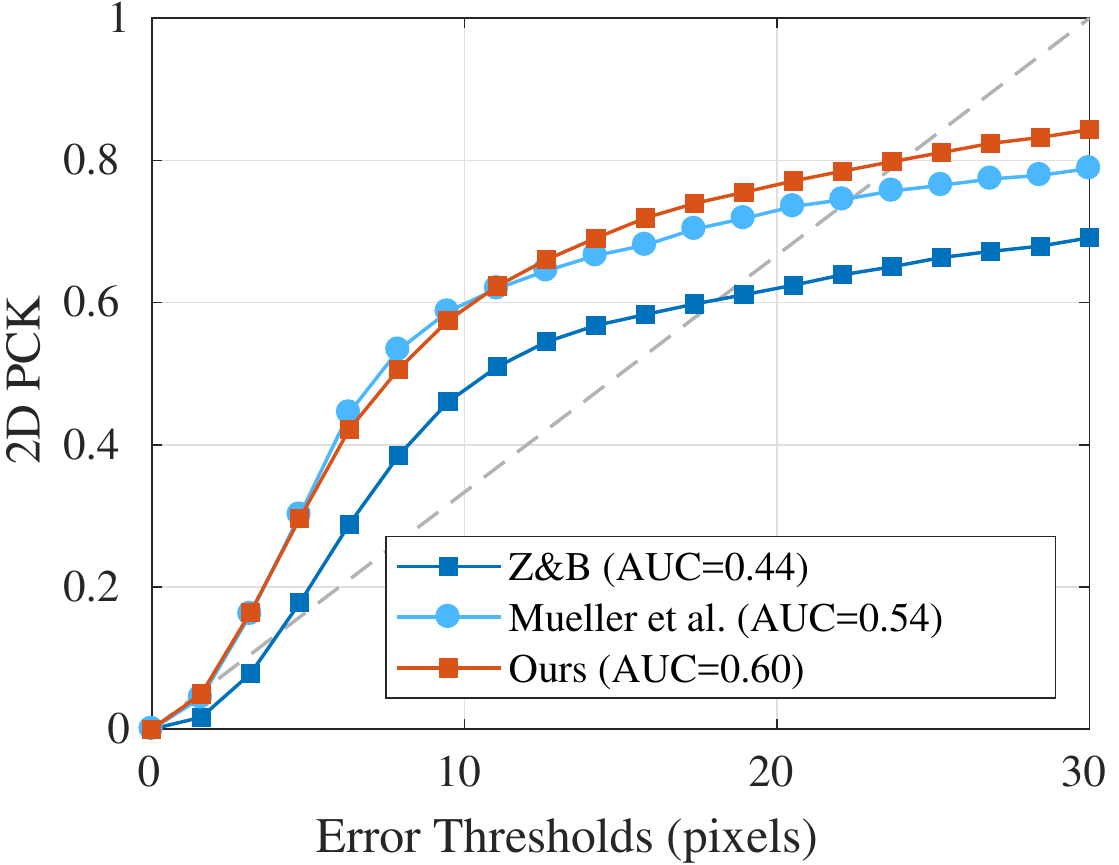}
	  \caption{2D PCK on ED dataset.}
	  \label{fig:ego_dexter_2d}
	\end{subfigure}
   \begin{subfigure}[t]{0.31\textwidth}
      \tiny
   	  \center
      \textbf{\tiny{EgoDexter}} 
	 \includegraphics[width=\columnwidth]{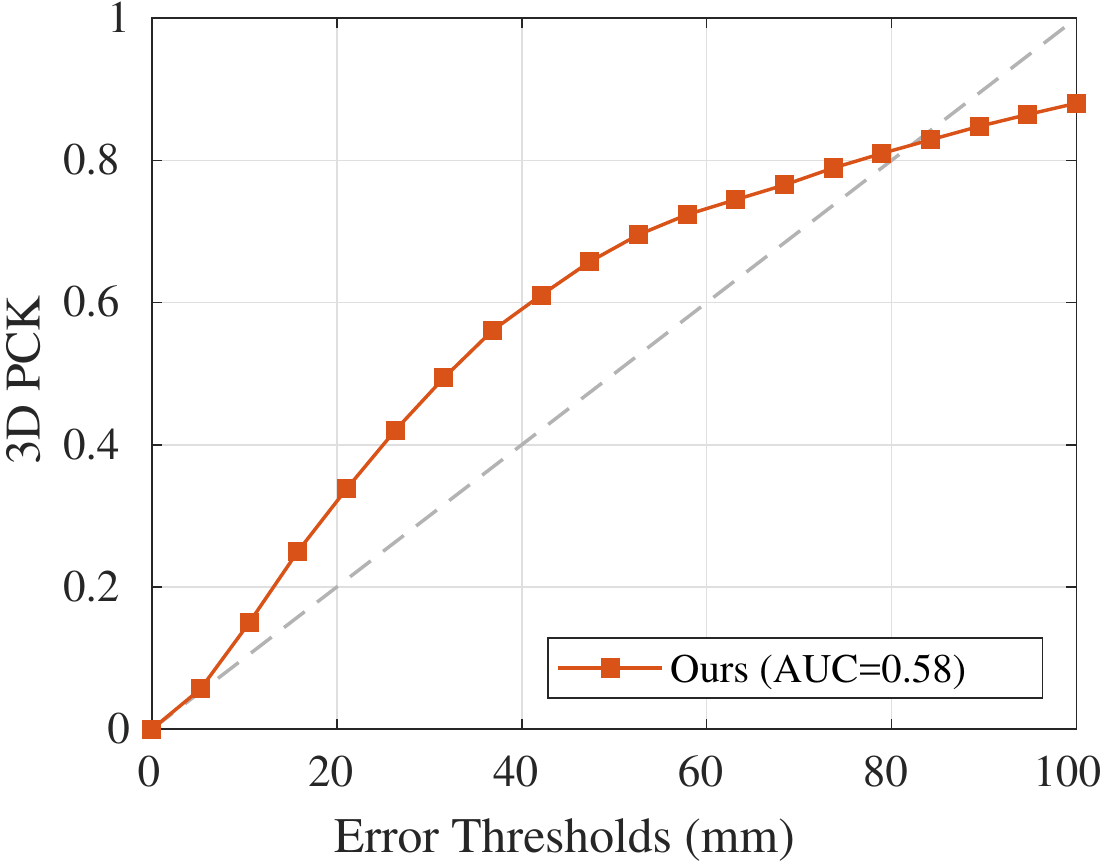}
	  \caption{3D PCK on ED dataset.} 
	  \label{fig:ego_dexter_3d}
	\end{subfigure}
    \begin{subfigure}[t]{0.31\textwidth}
     \center
    \textbf{\tiny{MPII+NZSL}} 
    \includegraphics[width=\columnwidth, height=0.75\columnwidth]{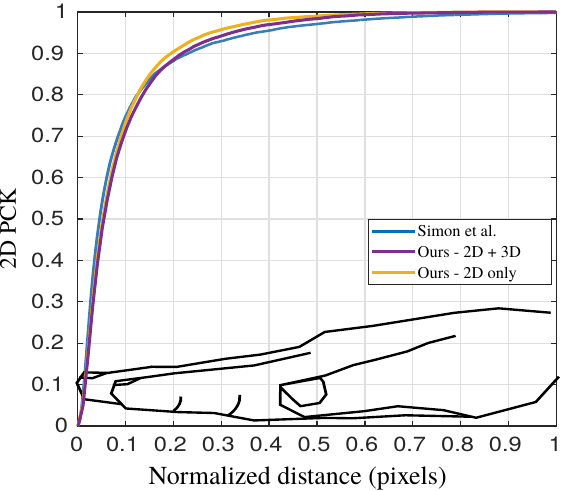}
     \caption{2D PCKh on MPII+NZSL. } 
  	\label{fig:mpii_nzsl_soa}
	\end{subfigure}
  \vspace{-2mm}
  \caption{Comparison with the state-of-the-art on the DO, ED, SHP and MPII+NZSL datasets.\vspace{-4mm}}
  \label{fig:quatitative_evaluation}
\end{figure*}

We provide a comparison of the proposed approach with state-of-the-art methods on all aforementioned datasets. Note that different approaches use different training data. We thus replicate the training setup of the corresponding approaches for a fair comparison. 

Fig.~\ref{fig:dexter_object_2d} and~\ref{fig:dexter_object_3d} compare the proposed approach with other methods on the D+O dataset for 2D and 3D pose estimation, respectively. In particular, we compare with the state-of-the-art approach by Zimmerman and Brox (Z\&B)~\cite{Zimmermann2017ICCV} and the contemporary work by Mueller~\etal~\cite{GANeratedHands_2018}. We use the same training data (SHP+RHP) for comparison with~\cite{Zimmermann2017ICCV} (AUC $0.64$ vs $0.49$), and only use additional data for comparison with~\cite{GANeratedHands_2018}(AUC $0.74$ vs $0.64$).
For the 3D pose estimation accuracy (Fig.~\ref{fig:dexter_object_3d}), the approach~\cite{Zimmermann2017ICCV} is not included since it only estimates scale normalized root-relative 3D pose.
Our approach clearly outperforms current RGB state-of-the-art method by Mueller~\etal~\cite{GANeratedHands_2018} by a large margin. 
The approach~\cite{GANeratedHands_2018} utilizes the video information to perform temporal smoothening and also performs subject specific adaptation under the assumption that the users hold their hand parallel to the camera image plane. In contrast, we only perform frame-wise predictions without temporal filtering or user assumptions.
Additionally, we report the results of the depth based approach by Sridhar \etal~\cite{sridhar_eccv2016}, which are obtained from~\cite{GANeratedHands_2018}. While the RGB-D sensor based approach~\cite{sridhar_eccv2016} still works better, our approach takes a giant leap forward as compared to the existing RGB based approaches. 

Fig.~\ref{fig:stereodataset} compares the proposed method with existing approaches on the SHP dataset. We use the same training data (SHP+RHP) as in \cite{Zimmermann2017ICCV} and outperform all existing methods despite the already saturated accuracy on the dataset and the additional training data and temporal information used in~\cite{GANeratedHands_2018}. 

Fig.~\ref{fig:ego_dexter_2d} compares the 2D pose estimation accuracy on the EgoDexter dataset. While we outperform all existing methods for 2D pose estimation, none of the existing approaches report their performance for 3D pose estimation on this dataset. We, however, also report our performance in Fig.~\ref{fig:ego_dexter_3d}. 

\begin{table}[t]
\centering
\scriptsize
\begin{tabularx}{\columnwidth}{Xccc|ccc}
\toprule
 \multirow{3}{6cm}{Method}  & \multicolumn{3}{c|}{2D Pose Estimation} &  \multicolumn{3}{c}{3D Pose Estimation} \\
 & \multirow{2}{1.2cm}{AUC $\uparrow$}    &  \multicolumn{2}{c|}{EPE (mm)}  &   \multirow{2}{1.2cm}{\centering AUC $\uparrow$}  &  \multicolumn{2}{c}{EPE (mm)}   \\
& & \multirow{1}{1.2cm}{median $\downarrow$ } & \multirow{1}{1.2cm}{mean $\downarrow$} & & \multirow{1}{1.2cm}{median  $\downarrow$} & \multirow{1}{1.2cm}{mean  $\downarrow$} \\
\midrule
Z \& B \cite{Zimmermann2017ICCV} & 0.72 & 5.00 & 9.14 & - & 18.8* & -\\
Ours  & 0.89 & 2.20 &  3.57 &  0.91 &  13.82 &  15.77 \\
\midrule
Ours w. GT $\hat{Z}_{root}$ and $\hat{s}$ & 0.89 & 2.20 &  3.57 &  0.94 &  11.33 &  13.41 \\ 
\bottomrule
\end{tabularx}
\vspace{-2mm}
\caption{Comparison with the state-of-the-art on the RHP dataset. *uses noisy ground-truth 2D poses for 3D pose estimation.\vspace{-5mm}}
\label{tab:rhd_soa}
\end{table}

The results on the RHP dataset are reported in Tab.~\ref{tab:rhd_soa}.  Our approach significantly outperforms~\cite{Zimmermann2017ICCV} even though they use ground-truth 2D poses to estimate the 3D poses. Since the dataset provides 3D pose annotations for complete hand skeleton, we also report the performance of the proposed approach when the ground-truth depth of the root joint and the global scale of the hand is known (w. GT $\hat{Z}_{root}$ and $\hat{s}$). We can see that our approach of 3D pose reconstruction and scale recovery is very close to the ground-truth. 

Finally, for completeness, in Fig.~\ref{fig:mpii_nzsl_soa} we compare our approach with~\cite{simon2017hand} which is a state-of-the-art approach for 2D pose estimation. The evaluation is performed on the test set of the MPII+NZSL dataset. We follow \cite{simon2017hand} and use the provided center location of the hand and the size of the head of the person to obtain the hand bounding box. 
We define a square bounding box with height and width equals to $0.7 \times head\textnormal{-}length$.
We report two variants of our method; 1) the model trained for both 2D and 3D pose estimation using the datasets for both tasks, and 2) a model trained for only 2D pose estimation using the same training data as in~\cite{simon2017hand}. In both cases we use the models trained with 2-stages. Our approach performs similar or better than~\cite{simon2017hand}, even though we use a smaller backbone network as compared to the 6-stage Convolutional Pose Machines (CPM) network \cite{wei2016convolutional} used in \cite{simon2017hand}. The CPM model with 6-stages has $51M$ parameters, while our $1$ and $2$-stage models have only $17M$ and $35M$ parameters, respectively. Additionally, our approach also infers the 3D hand pose. 

\begin{figure}[t]
  \centering
  \setlength{\tabcolsep}{5pt}
\scalebox{1}{
  \begin{tabular}{c c c c c c c}

	& \tiny{2D Pose} & \tiny{3D Pose} & \tiny{2D Pose} & \tiny{3D Pose}	& \tiny{2D Pose} & \tiny{3D Pose}\\
 	\rotatebox{90}{\centering \tiny{Dexter+Object}} &
    \includegraphics[trim={1.7cm 1cm 1.5cm .5cm},clip,height=0.130\linewidth]{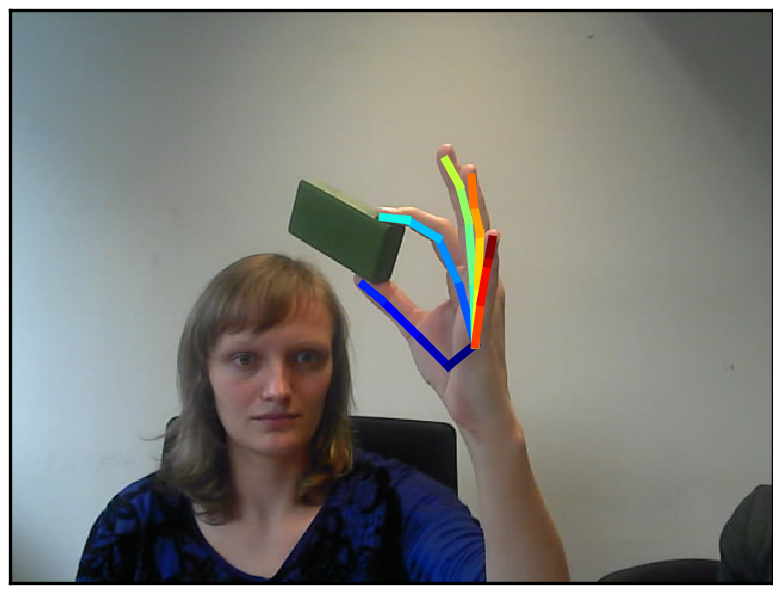}   &
    \includegraphics[trim={9cm 2cm 9cm 0cm},clip,height=0.13\linewidth]{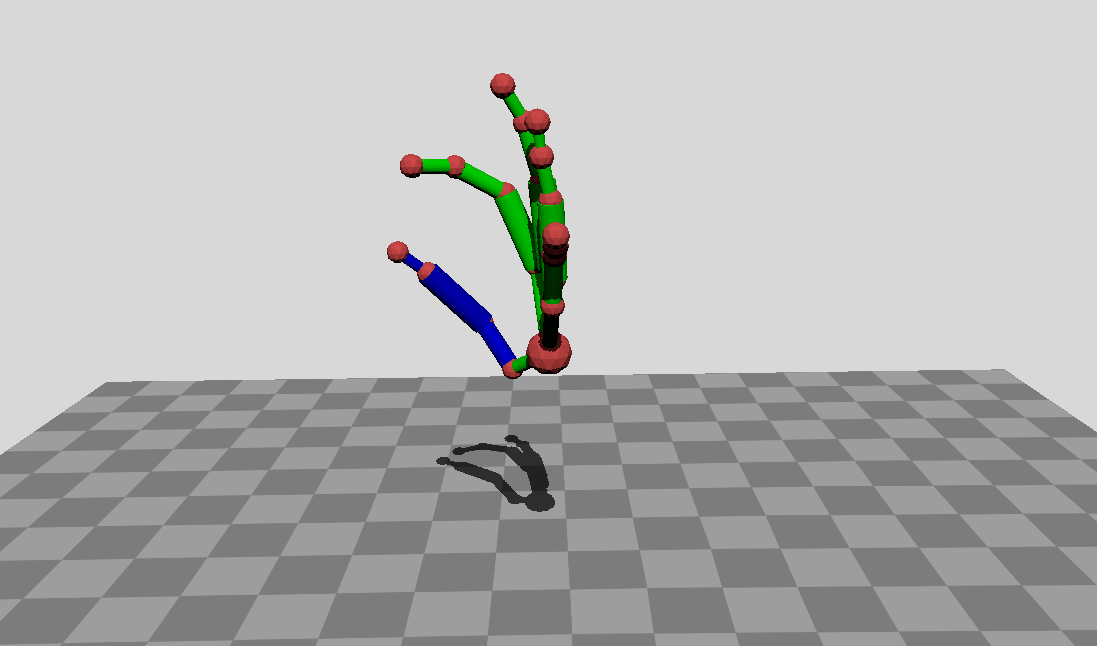}  &    
    \includegraphics[trim={1.6cm 1cm 1.55cm .5cm},clip,height=0.130\linewidth]{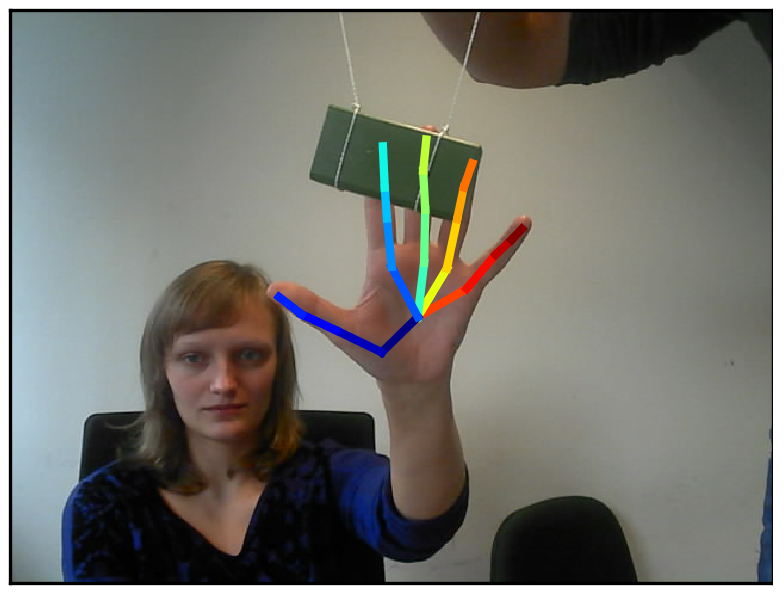} &  
    \includegraphics[trim={9cm 2cm 9cm 0cm},clip,height=0.130\linewidth]{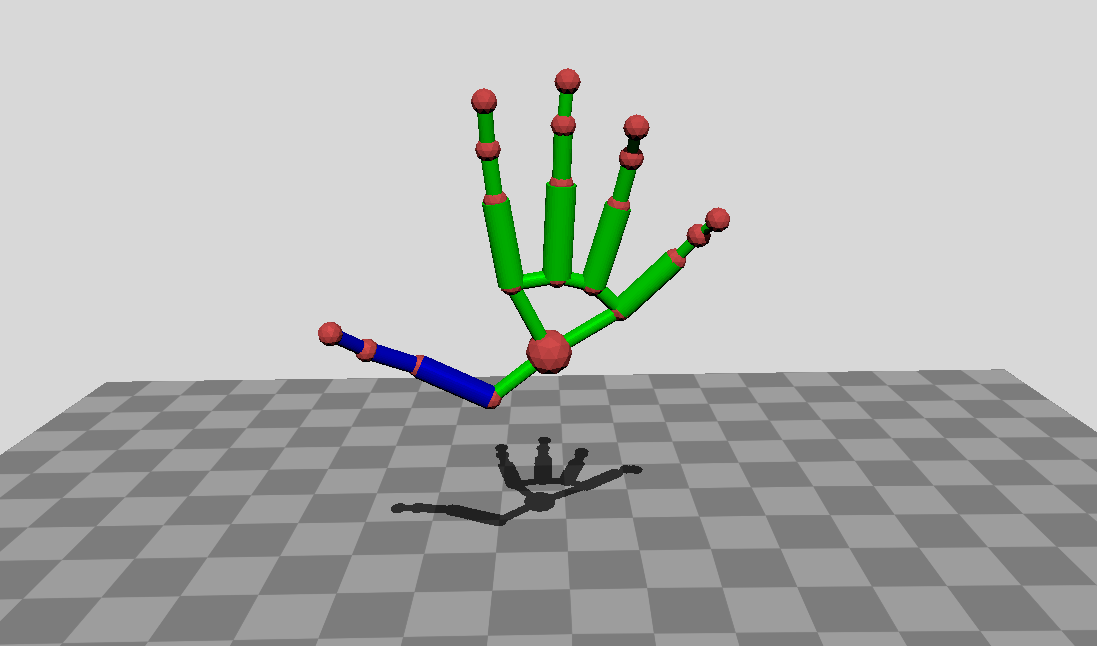} &
    \includegraphics[trim={1.4cm 0.16cm 1.4cm 1cm},clip,height=0.130\linewidth]{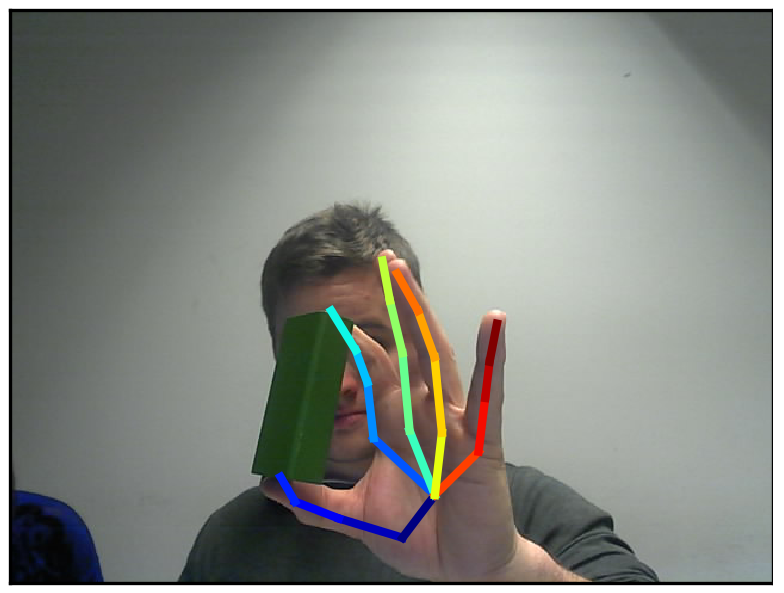}   &
    \includegraphics[trim={9cm 2cm 9cm 0cm},clip,height=0.130\linewidth]{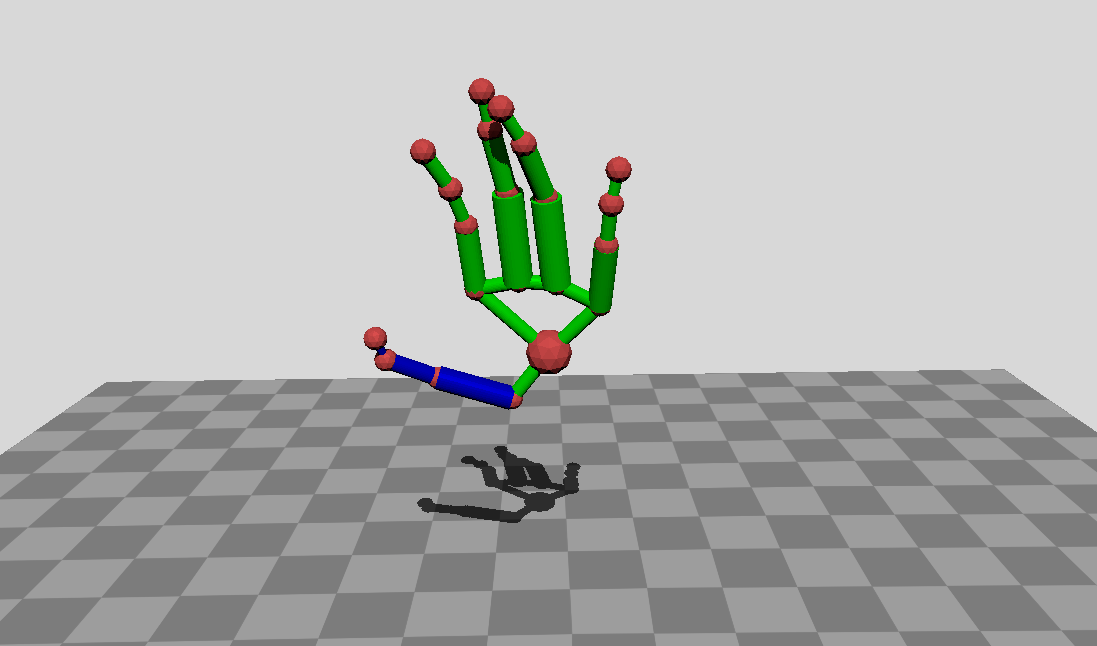} 
    \vspace{-1mm}\\
    \rotatebox{90}{\centering \tiny{~~EgoDexter}} &
    \includegraphics[trim={0.6cm .0cm 1.2cm .15cm},clip,height=0.130\linewidth]{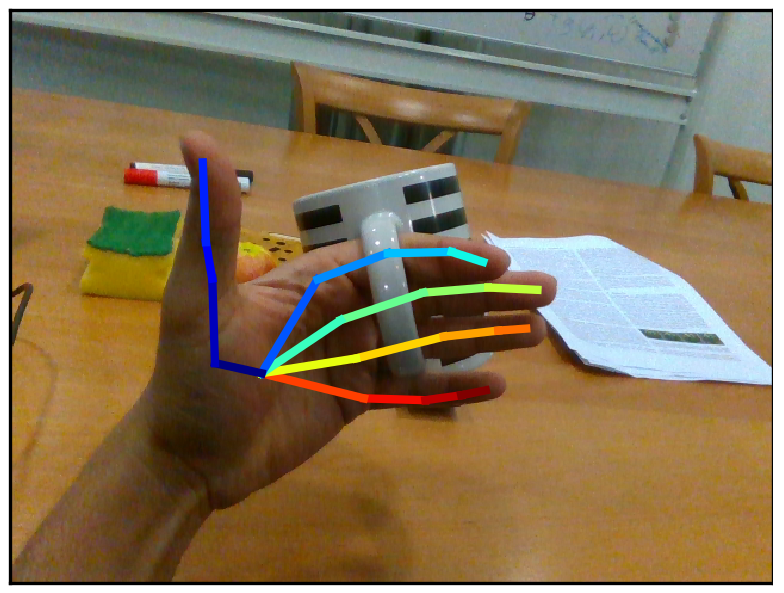}   &
    \includegraphics[trim={9cm 2cm 9cm 0cm},clip,height=0.130\linewidth]{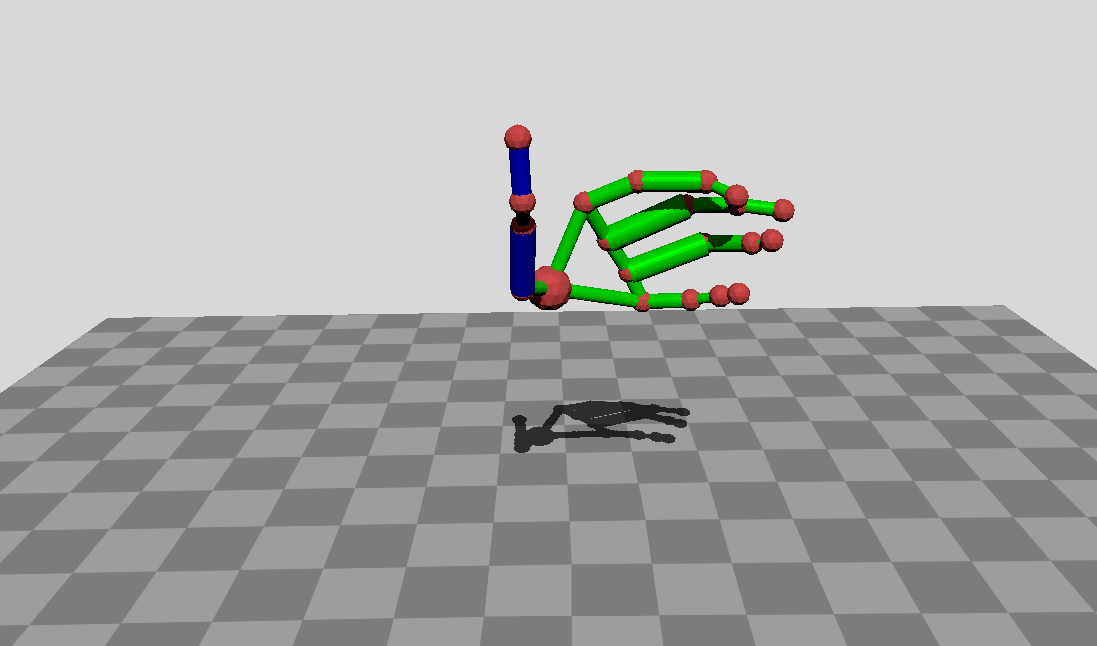}  &    
    \includegraphics[trim={1.3cm 1cm 1.4cm 0.1cm},clip,height=0.130\linewidth]{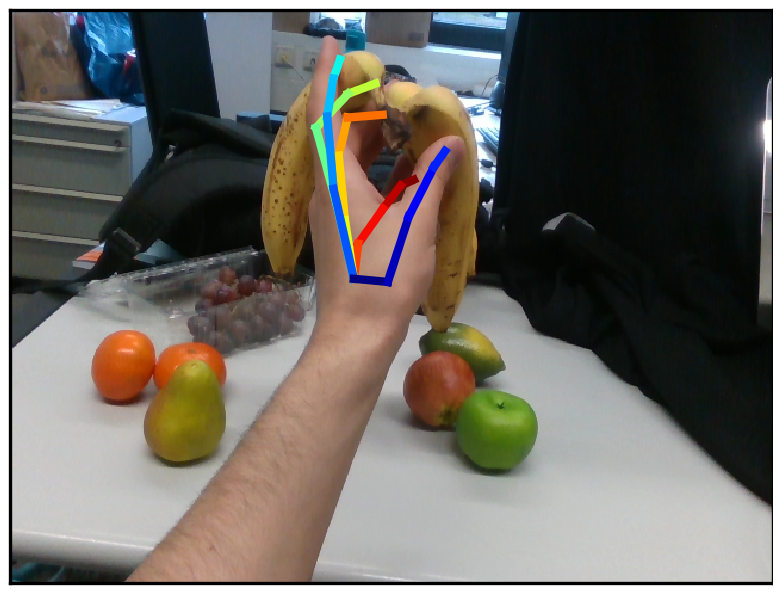} &  
    \includegraphics[trim={9cm 2cm 9cm 0cm},clip,height=0.130\linewidth]{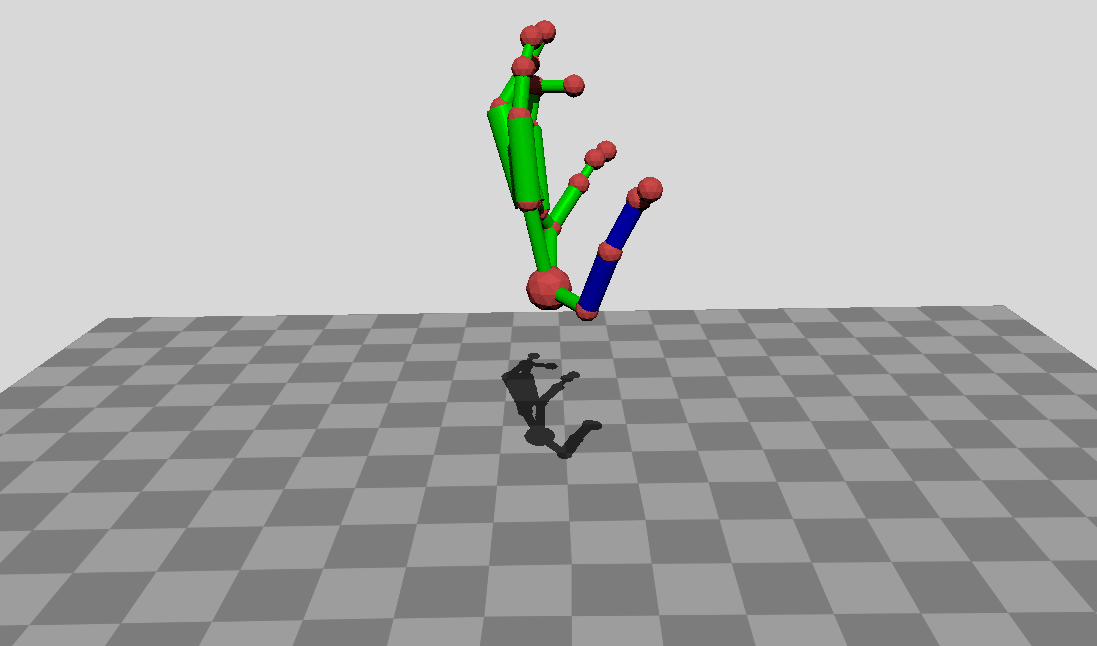} &
    \includegraphics[trim={.25cm 1cm 2.5cm 0.13cm},clip,height=0.130\linewidth]{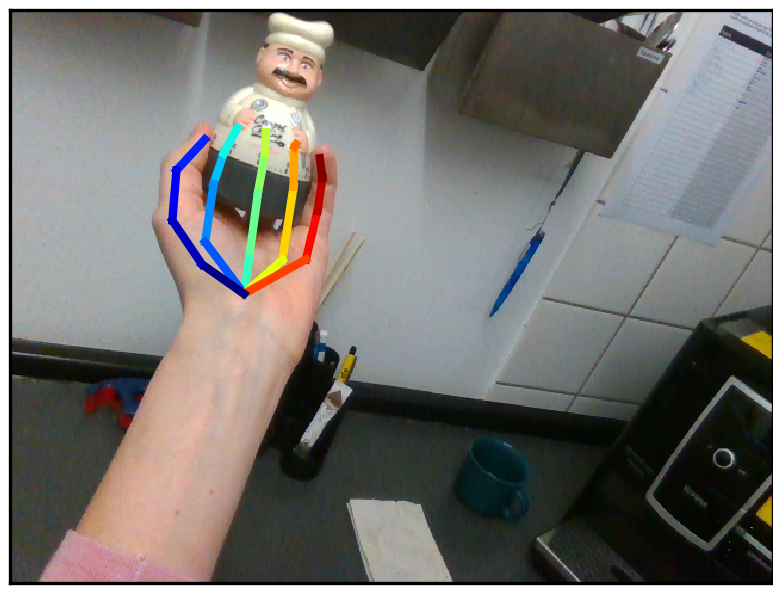}   &
    \includegraphics[trim={9cm 2cm 9cm 0cm},clip,height=0.130\linewidth]{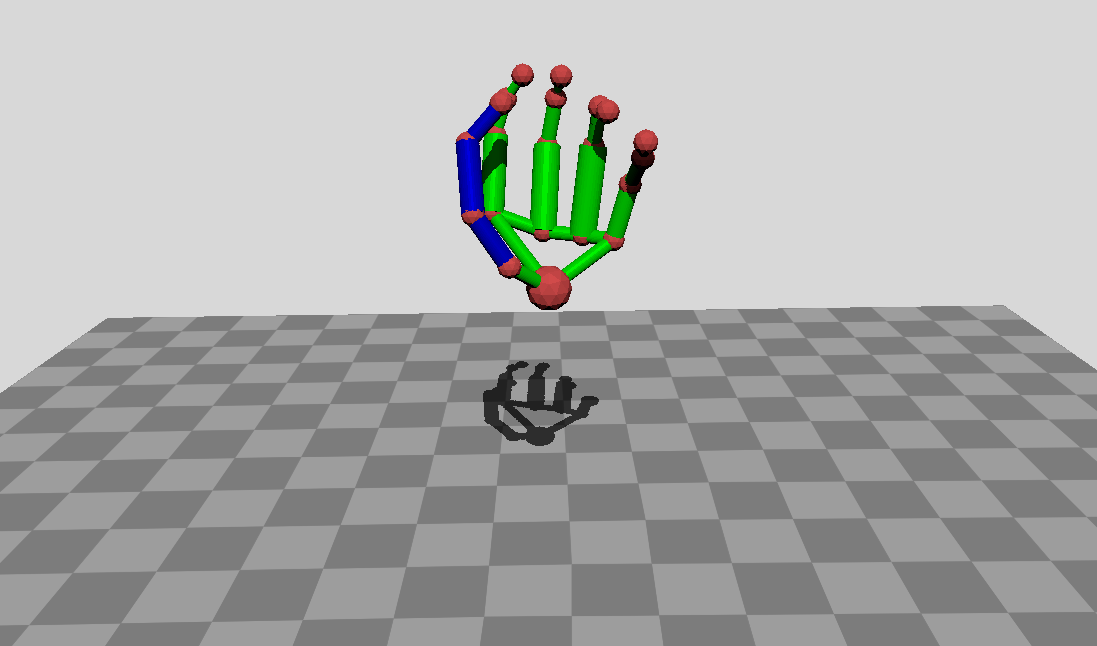} 
    \vspace{-1mm}\\
  	\rotatebox{90}{\centering \tiny{~~~~~~~SHP}} &
    \includegraphics[trim={1.2cm .5cm 1.9cm 1cm},clip,height=0.130\linewidth]{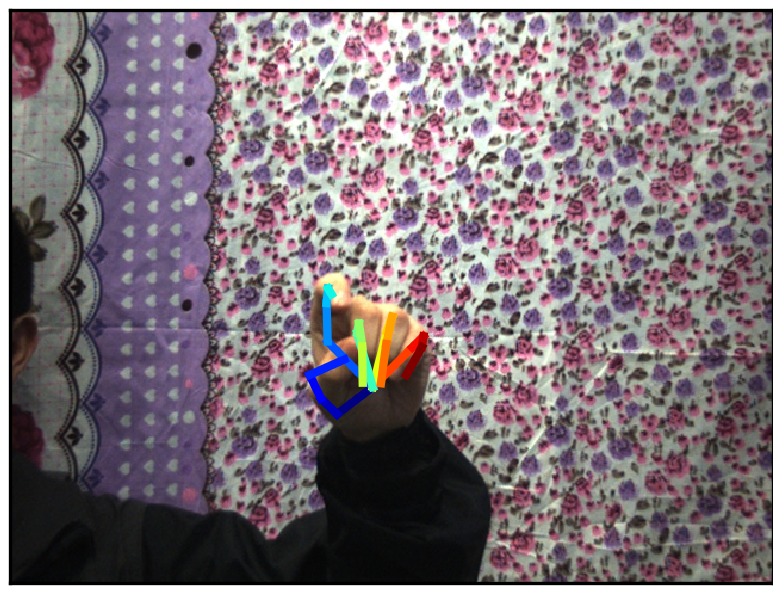} &  
    \includegraphics[trim={9cm 2cm 9cm 0cm},clip,height=0.130\linewidth]{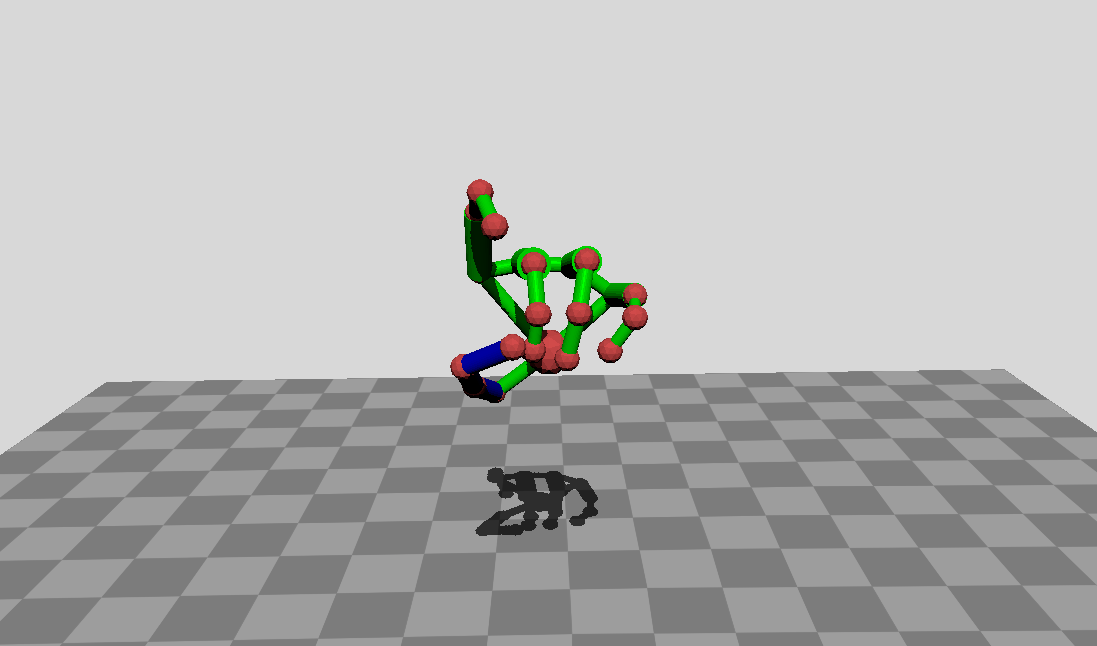} &
    \includegraphics[trim={1.1cm .5cm 2cm 1cm},clip,height=0.130\linewidth]{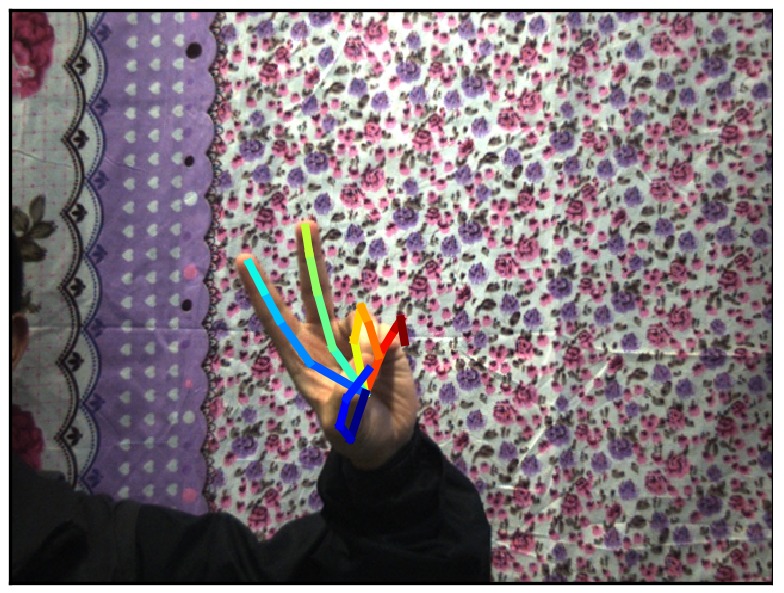}   &
    \includegraphics[trim={9cm 2cm 9cm 0cm},clip,height=0.130\linewidth]{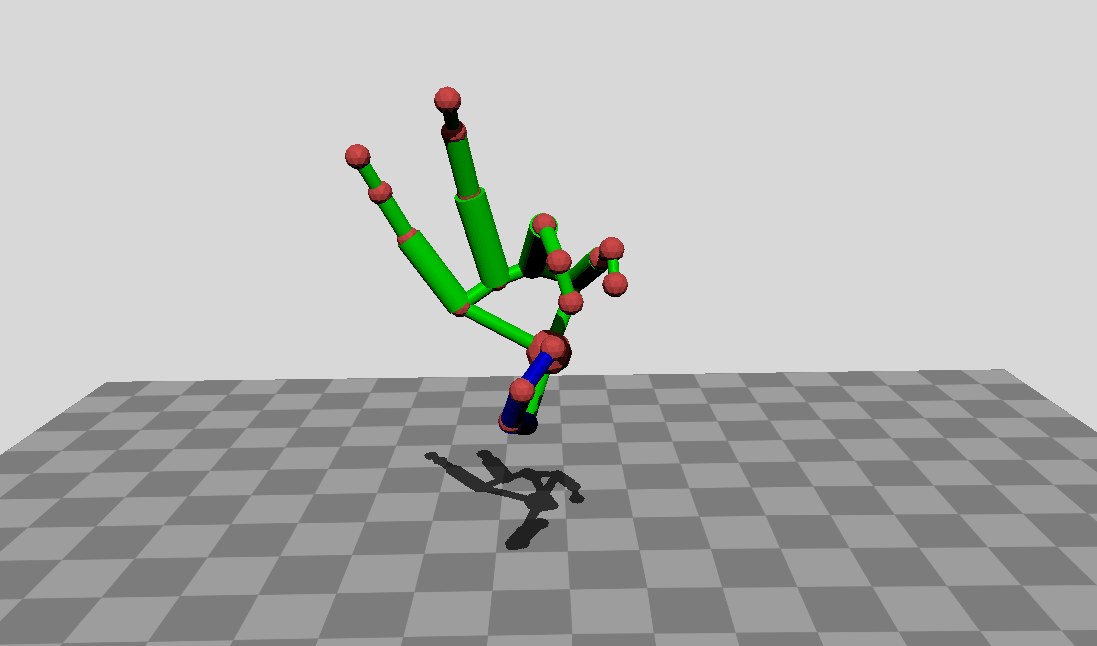} &
    \includegraphics[trim={1.1cm .5cm 2cm 1cm},clip,height=0.130\linewidth]{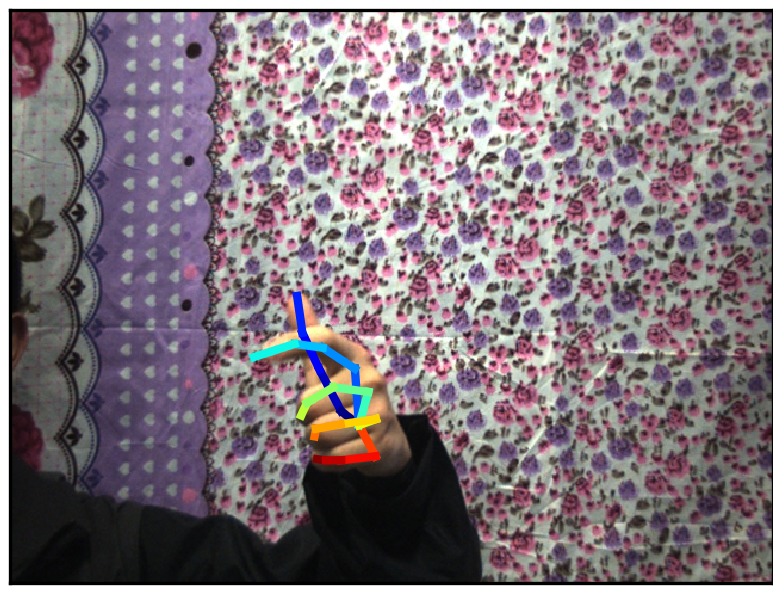}   &
    \includegraphics[trim={9cm 2cm 9cm 0cm},clip,height=0.130\linewidth]{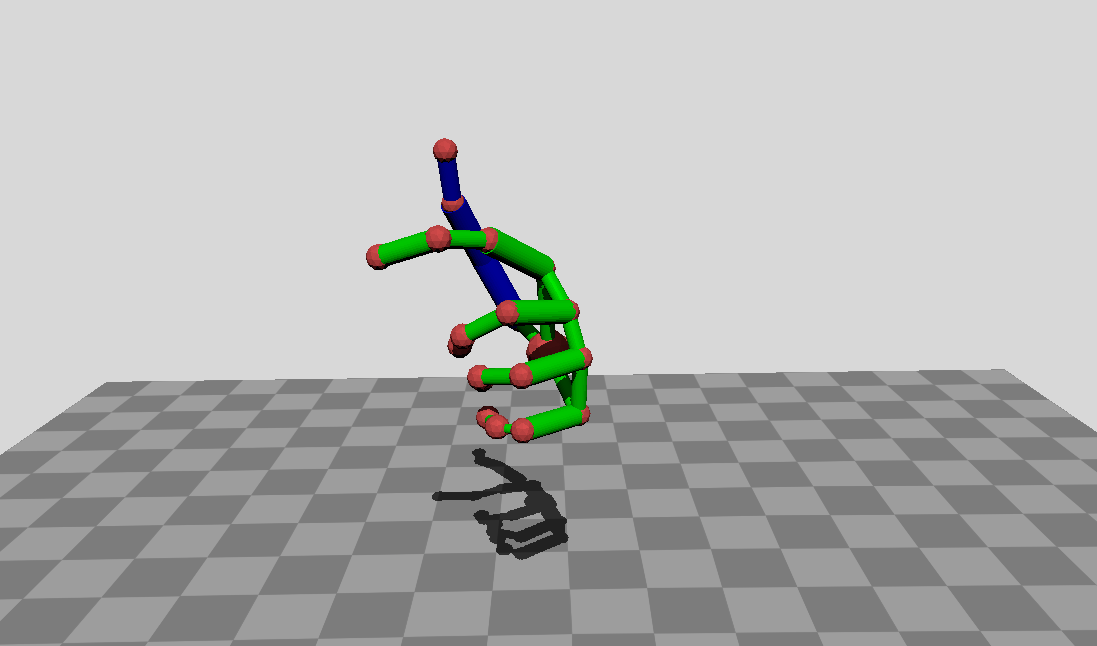}  
  	\vspace{-1mm}\\
  	\rotatebox{90}{\centering \tiny{~~~~~~MPII}} &
    \includegraphics[trim={1cm 1cm 1cm 1cm},clip,height=0.130\linewidth]{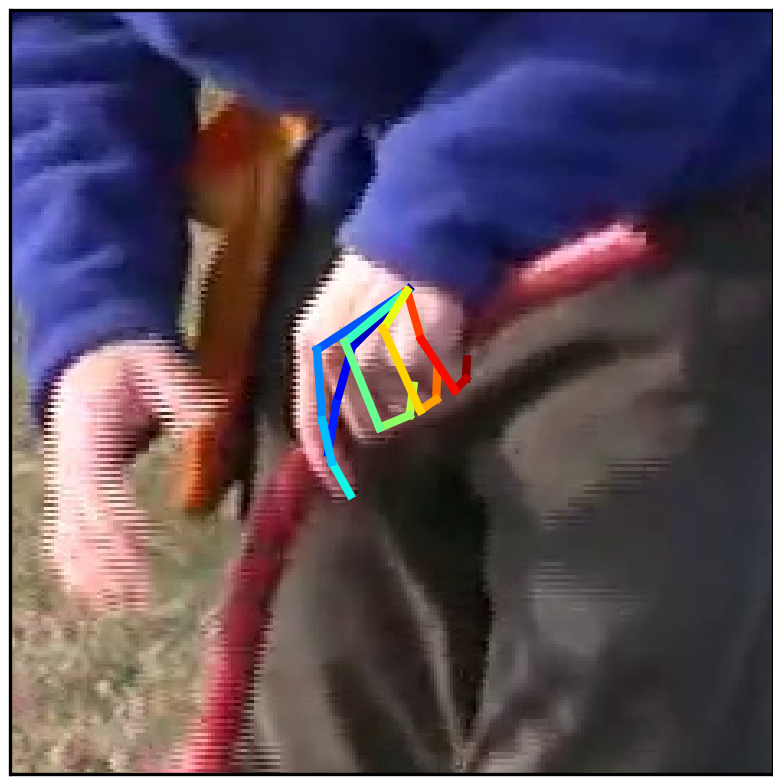} &  
    \includegraphics[trim={10cm 2cm 10cm 2cm},clip,height=0.130\linewidth]{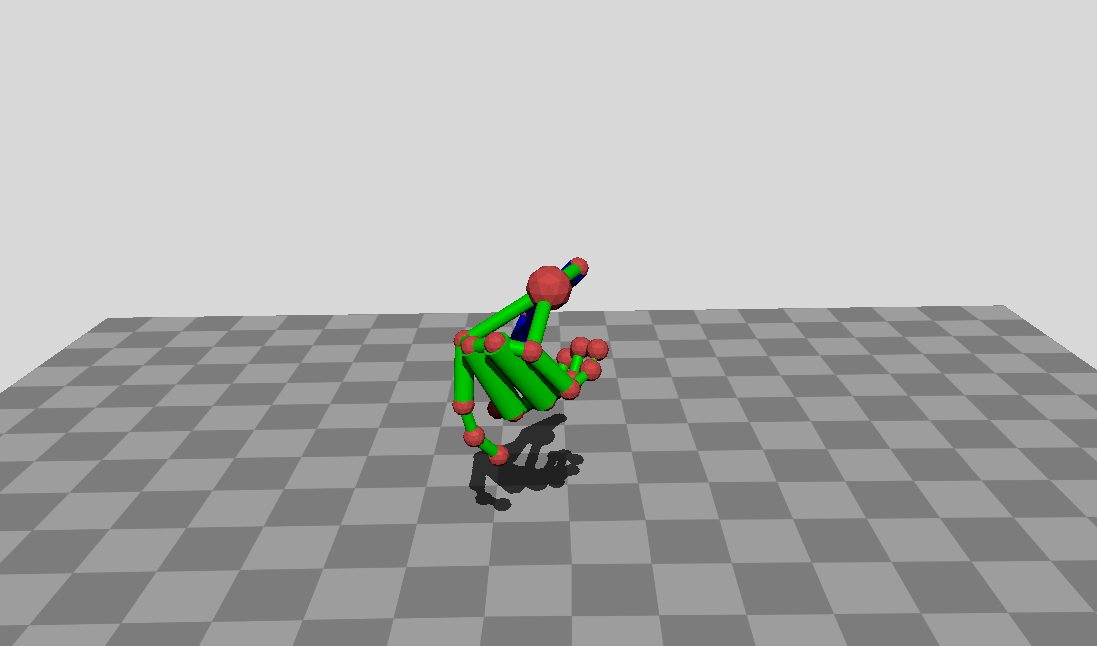} &
    \includegraphics[trim={1cm 1cm 1cm 1cm},clip,height=0.130\linewidth]{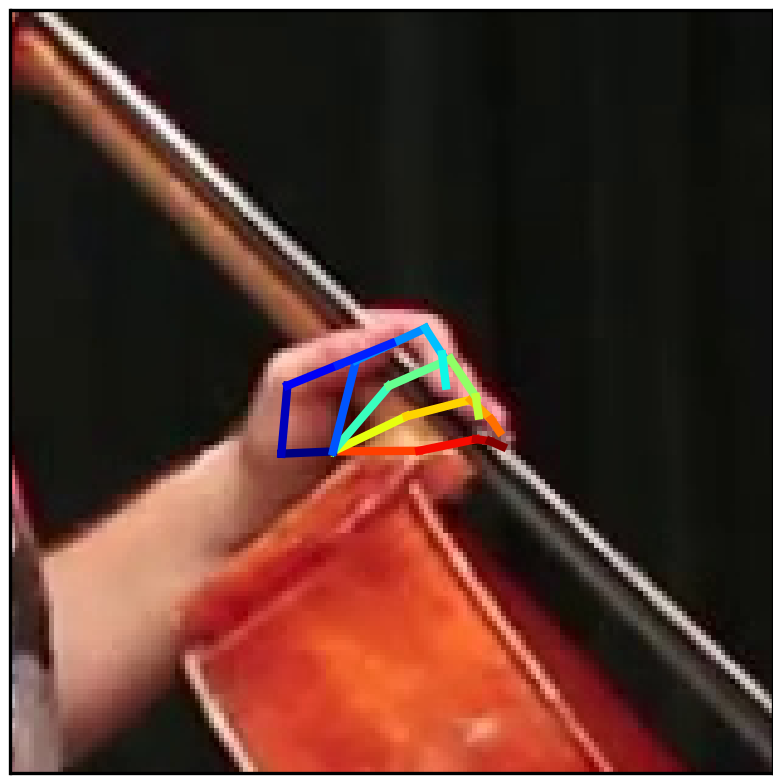}   &
    \includegraphics[trim={10cm 2cm 10cm 2cm},clip,height=0.130\linewidth]{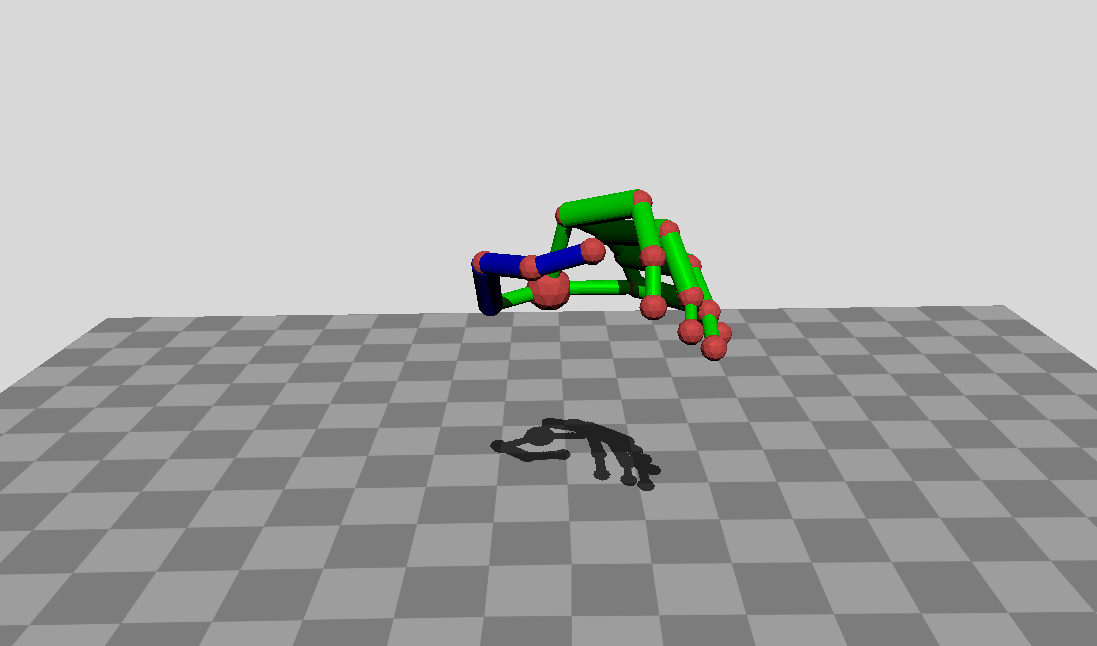} &
    \includegraphics[trim={1cm 1cm 1cm 1cm},clip,height=0.130\linewidth]{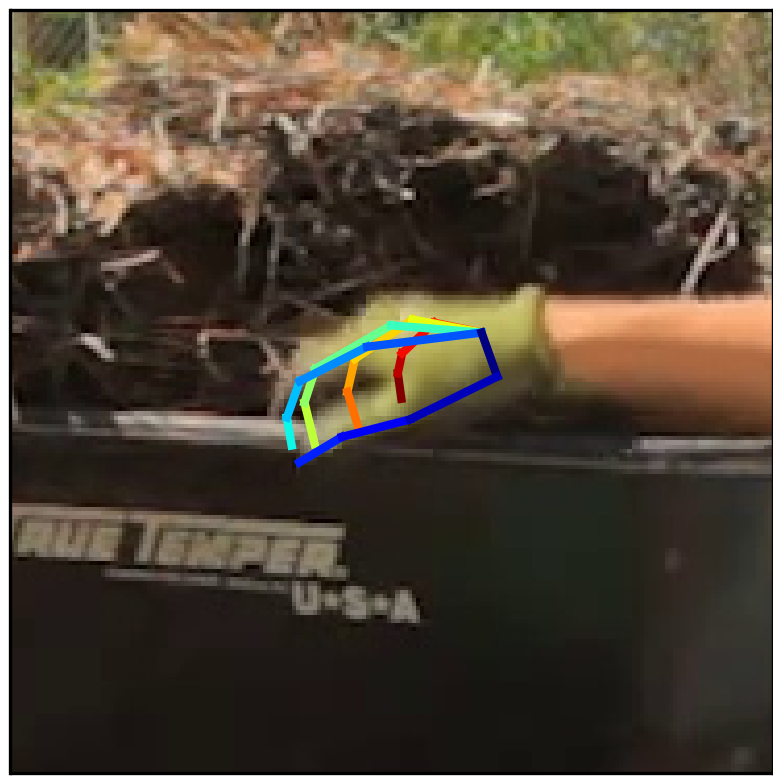}   &
    \includegraphics[trim={10cm 2cm 10cm 2cm},clip,height=0.130\linewidth]{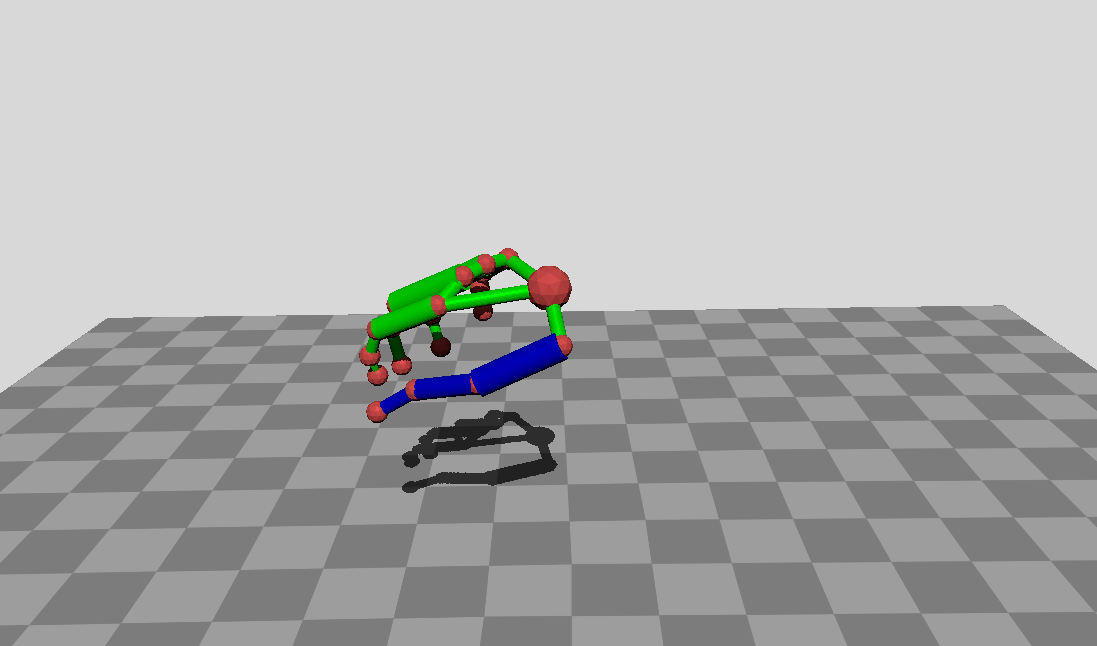}
    \vspace{-1mm} \\
  	\rotatebox{90}{\centering \tiny{~~~~~~MPII}} &
    \includegraphics[trim={1cm 1cm 1cm 1cm},clip,height=0.130\linewidth]{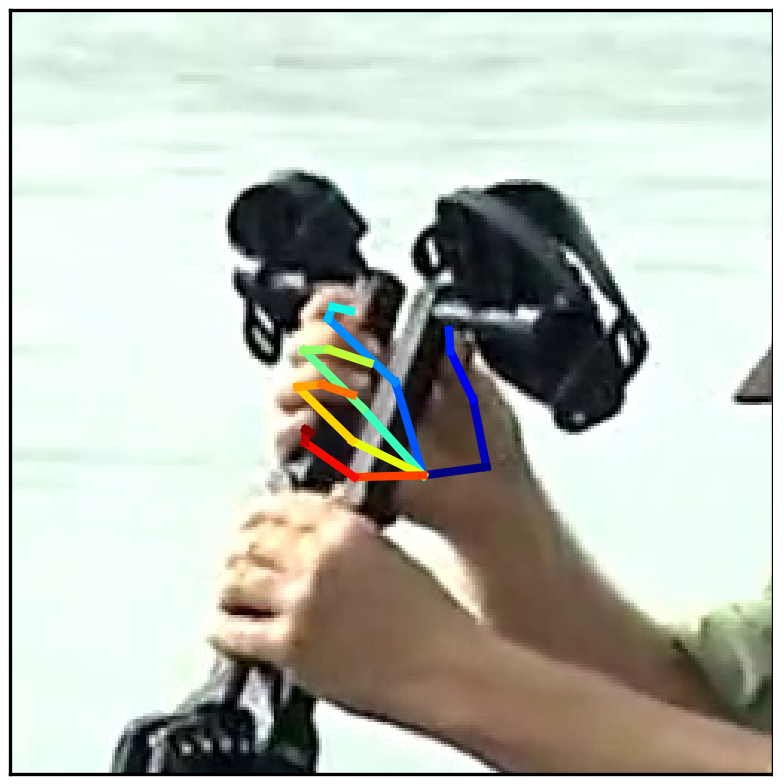} &  
    \includegraphics[trim={10cm 2cm 10cm 2cm},clip,height=0.130\linewidth]{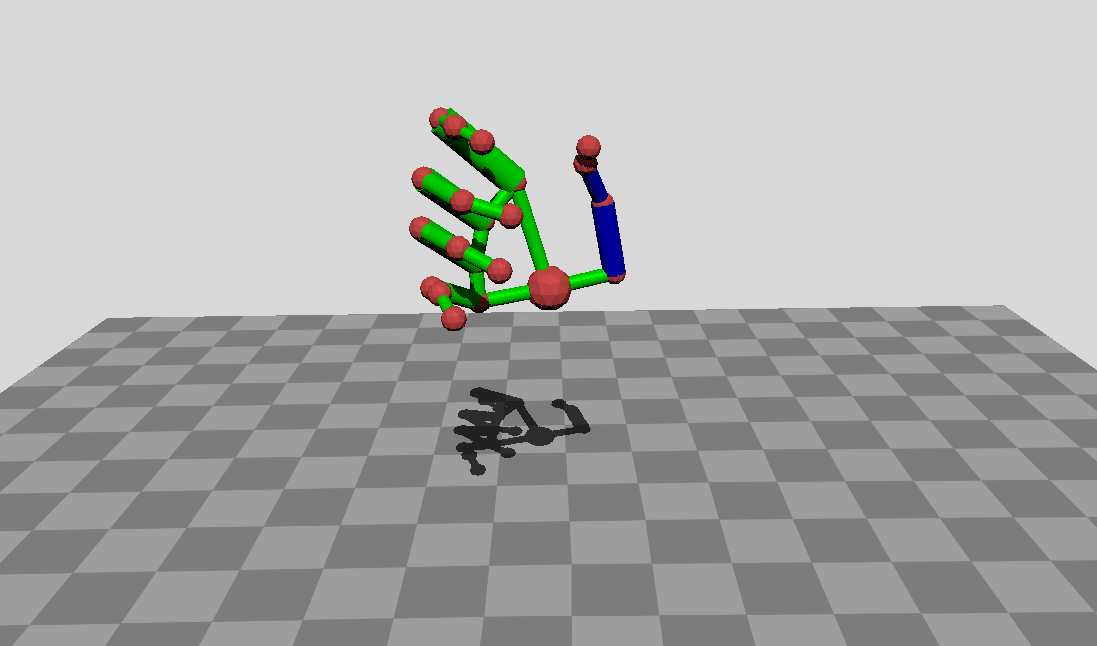} &
    \includegraphics[trim={1cm 1cm 1cm 1cm},clip,height=0.130\linewidth]{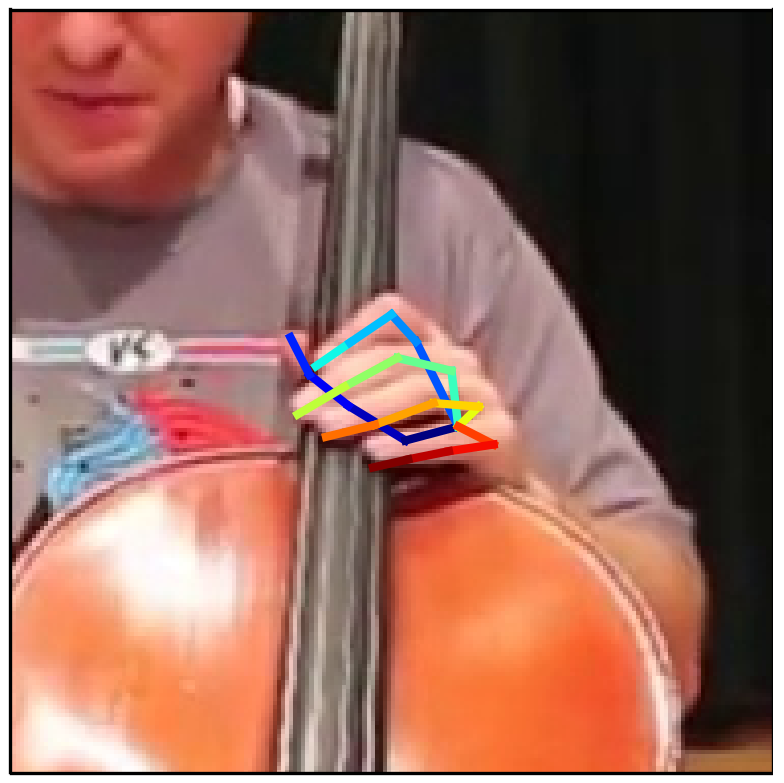}   &
    \includegraphics[trim={10cm 2cm 10cm 2cm},clip,height=0.130\linewidth]{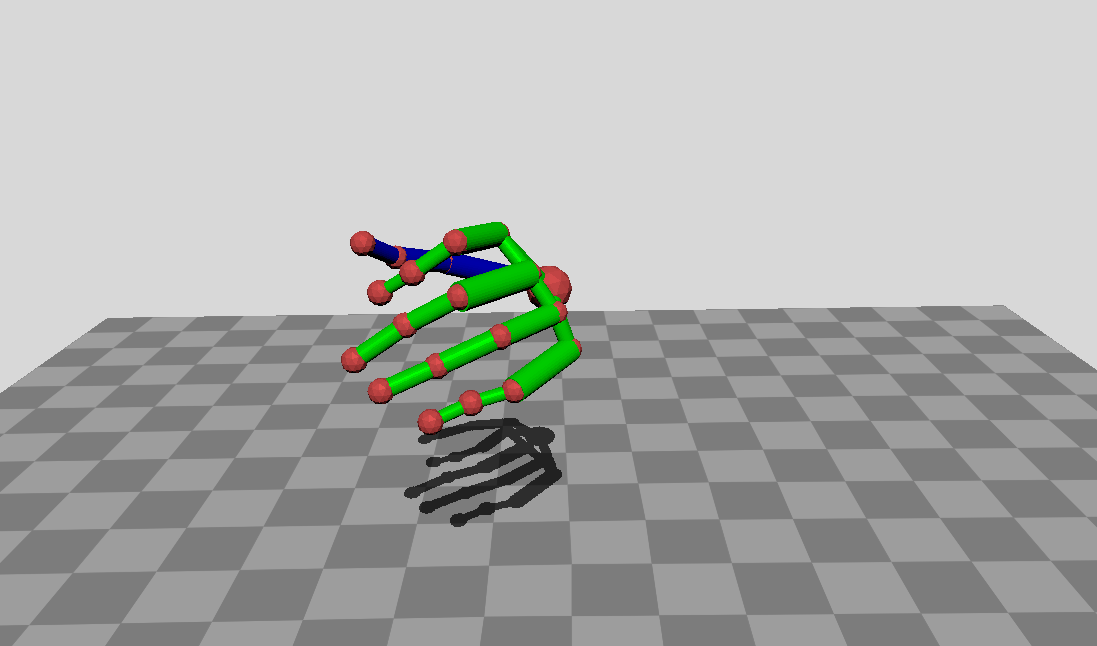} &
    \includegraphics[trim={1cm 1cm 1cm 1cm},clip,height=0.130\linewidth]{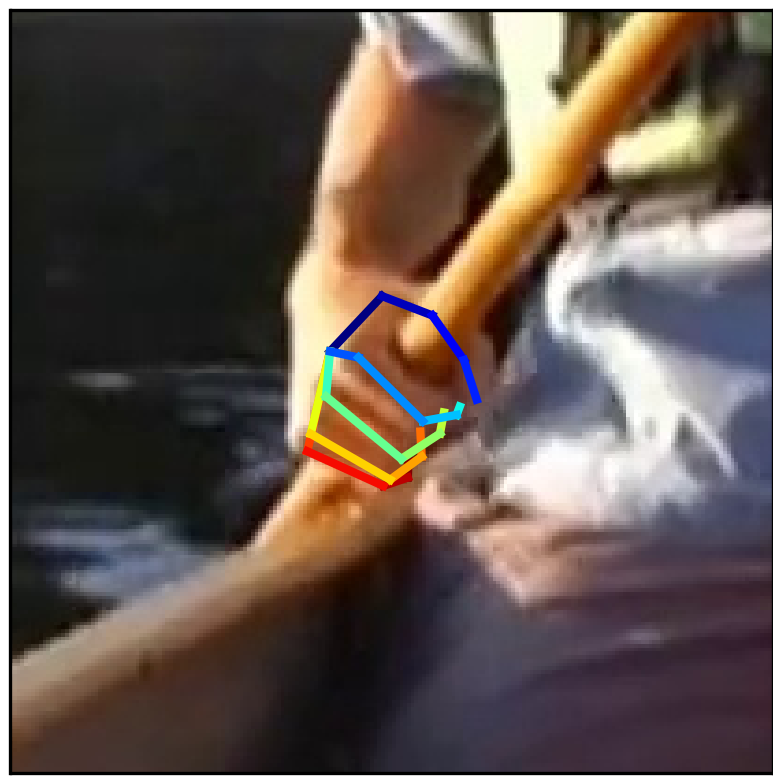}   &
    \includegraphics[trim={10cm 2cm 10cm 2cm},clip,height=0.130\linewidth]{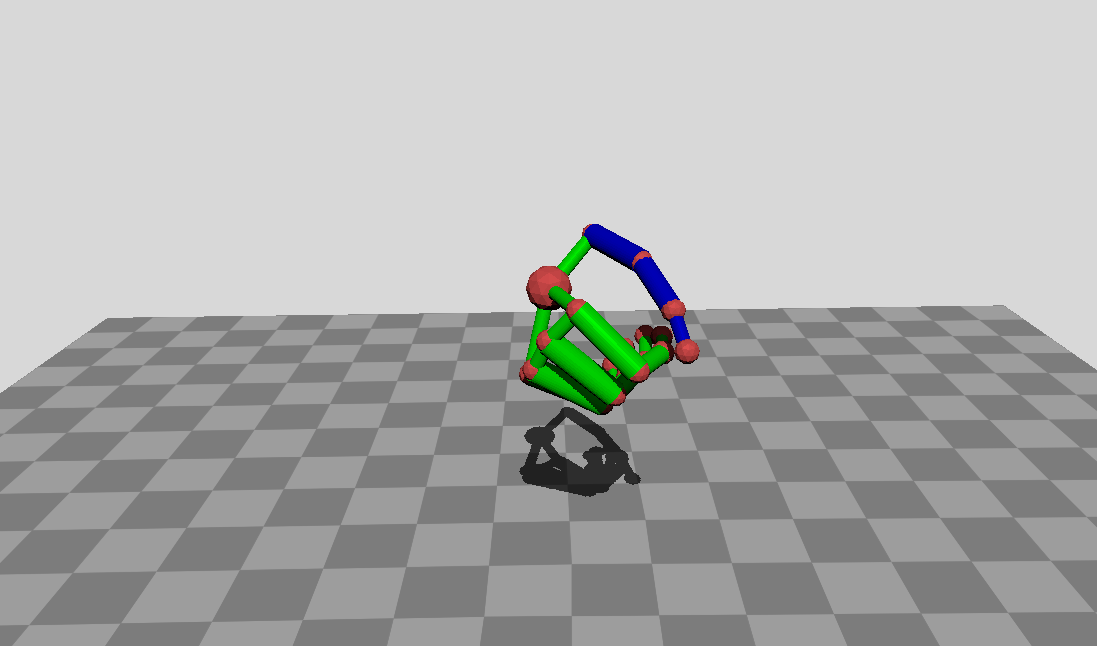}
    \vspace{-1mm}\\
 	\rotatebox{90}{\centering \tiny{~~~~~~MPII}} &
   \includegraphics[trim={1cm 1cm 1cm 1cm},clip,height=0.130\linewidth]{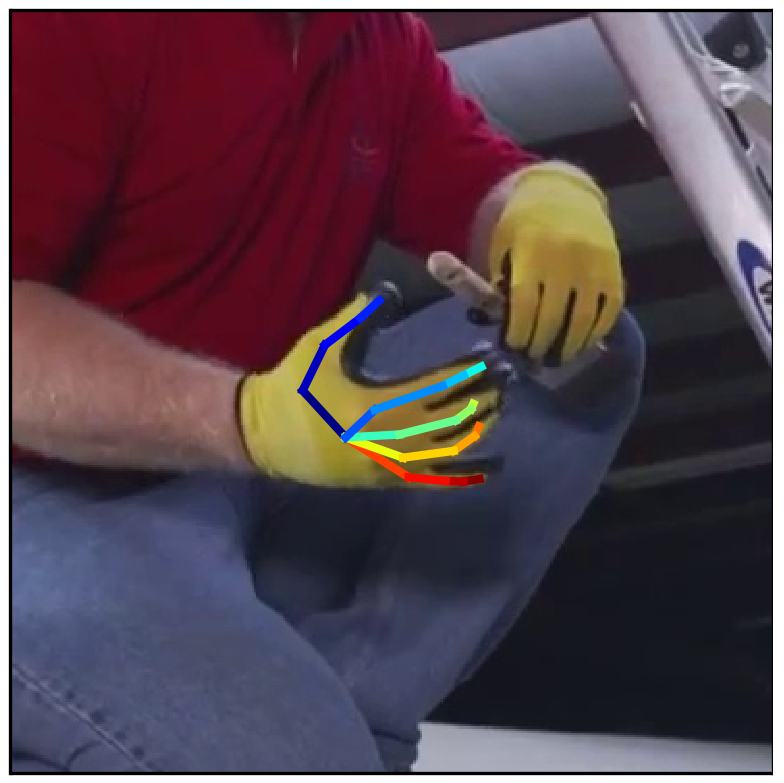} &  
   \includegraphics[trim={10cm 2cm 10cm 2cm},clip,height=0.130\linewidth]{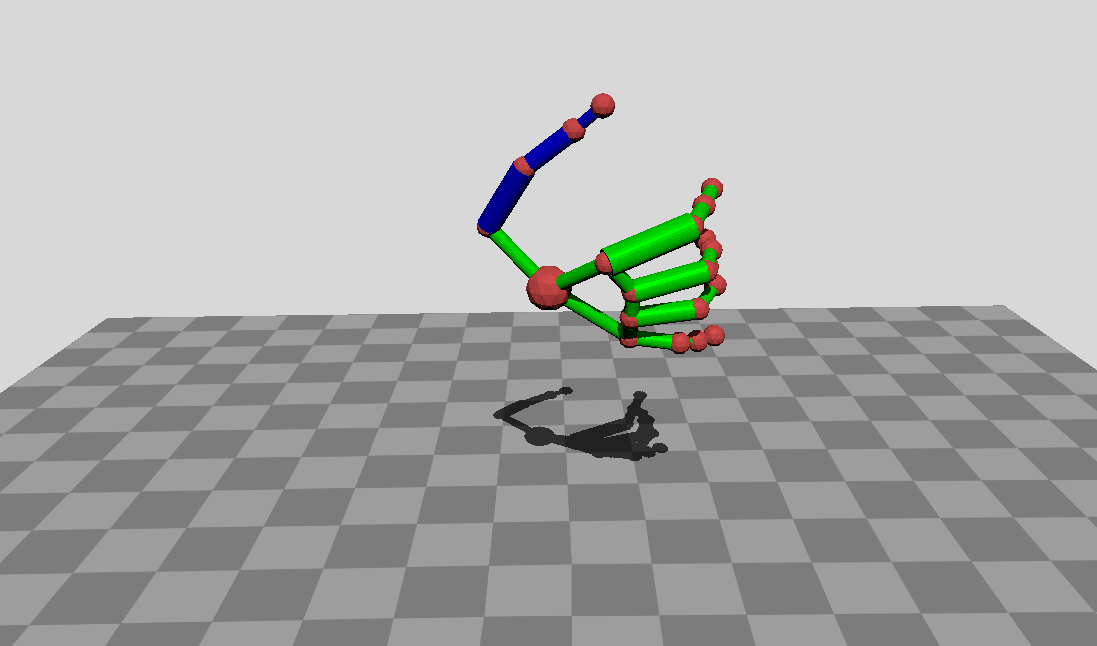} &
   \includegraphics[trim={1cm 1cm 1cm 1cm},clip,height=0.130\linewidth]{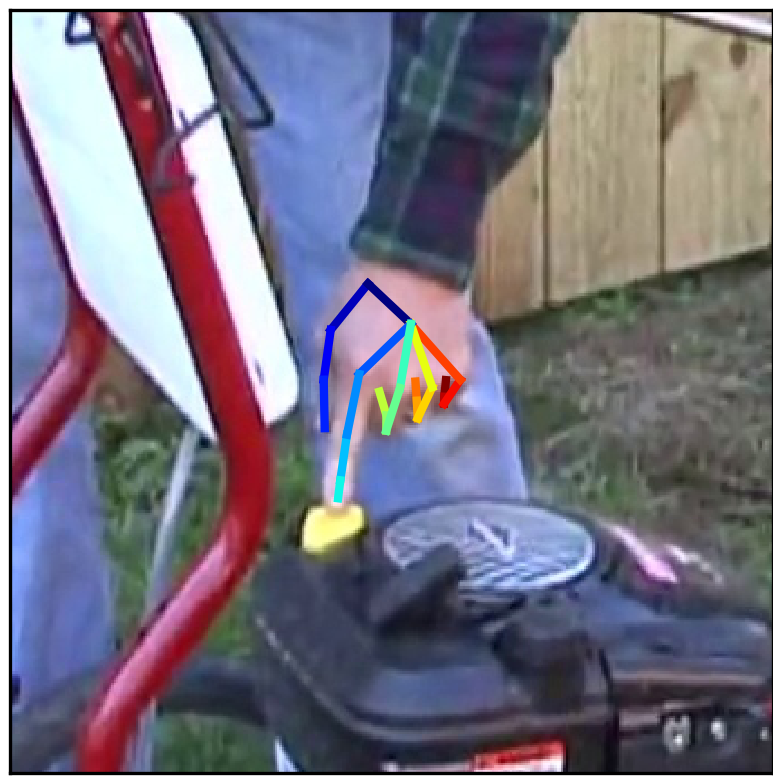}   &
   \includegraphics[trim={10cm 2cm 10cm 2cm},clip,height=0.130\linewidth]{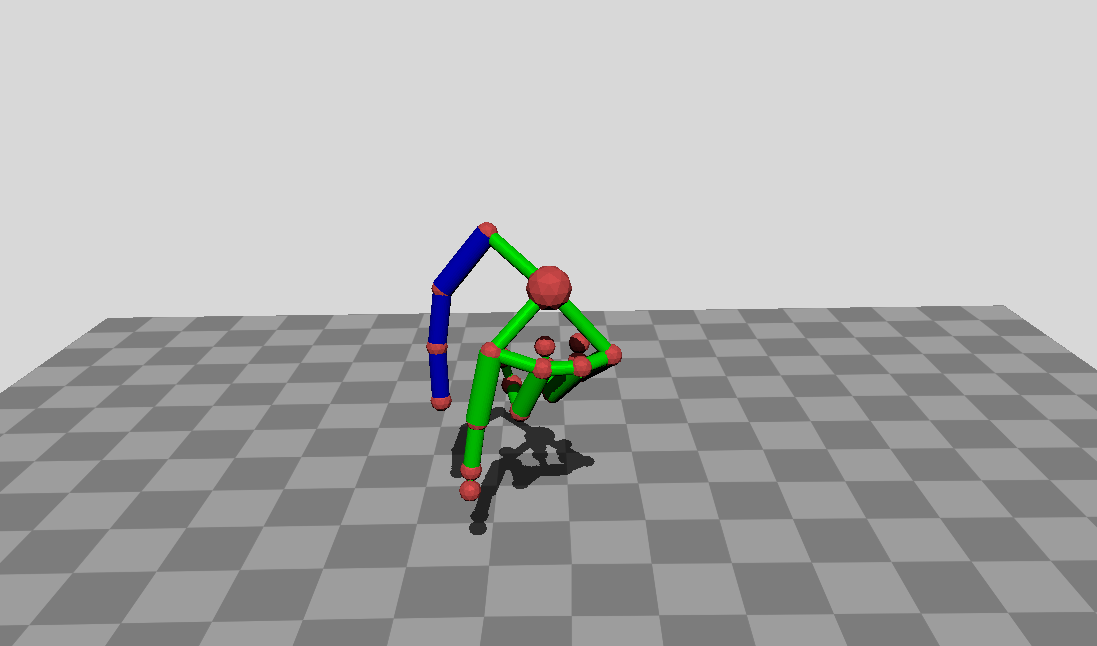} &
   \includegraphics[trim={1cm 1cm 1cm 1cm},clip,height=0.130\linewidth]{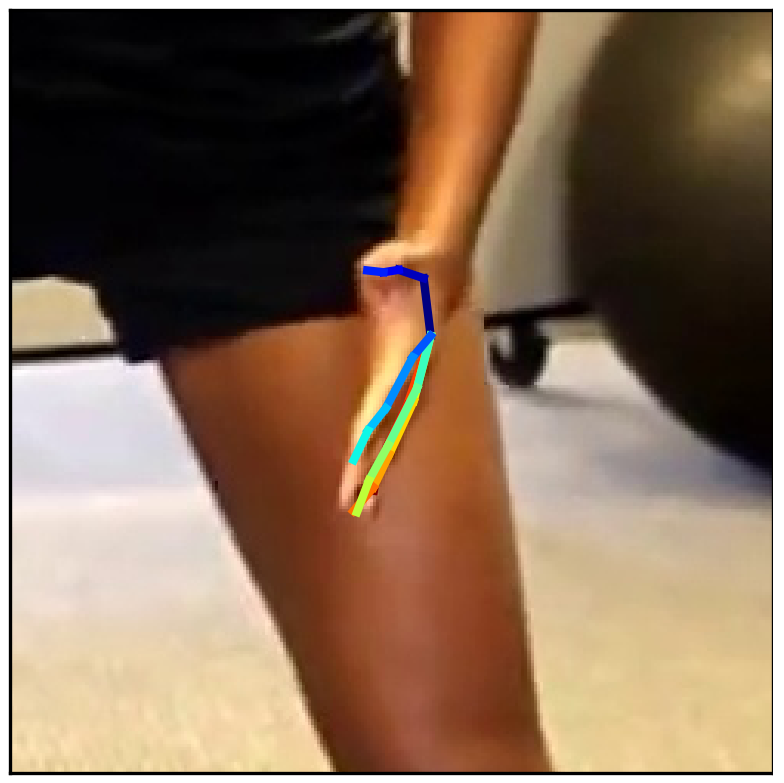}   &
   \includegraphics[trim={10cm 2cm 10cm 2cm},clip,height=0.130\linewidth]{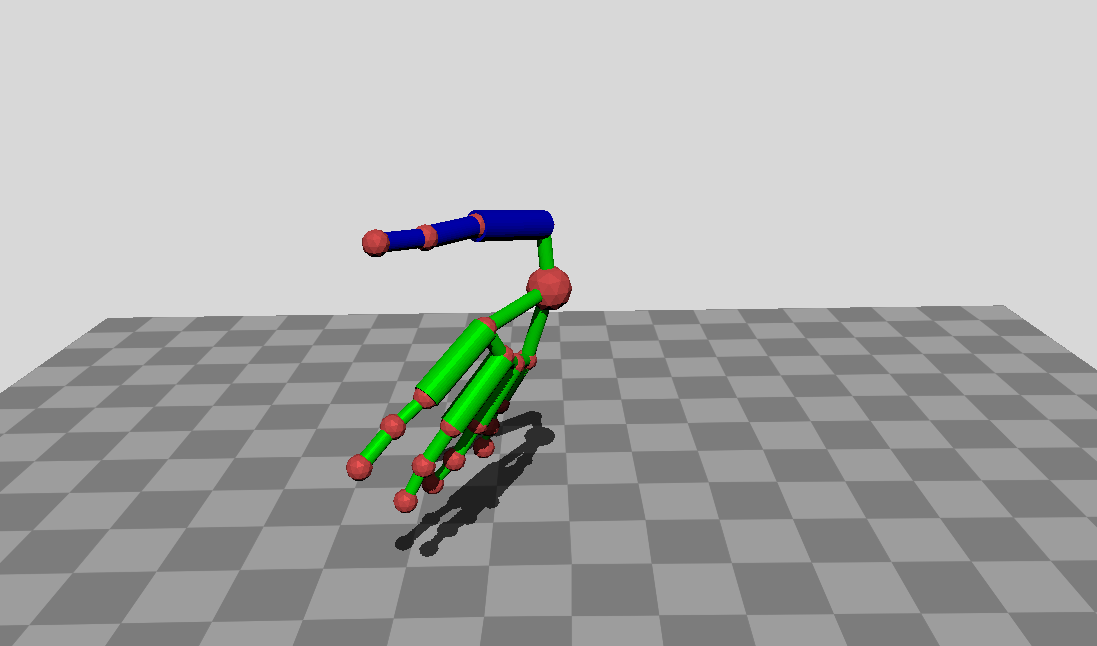}
   \vspace{-1mm} \\
  	\rotatebox{90}{\centering \tiny{~~~~~~NZSL}} &
    \includegraphics[trim={1cm 1cm 1cm 1cm},clip,height=0.130\linewidth]{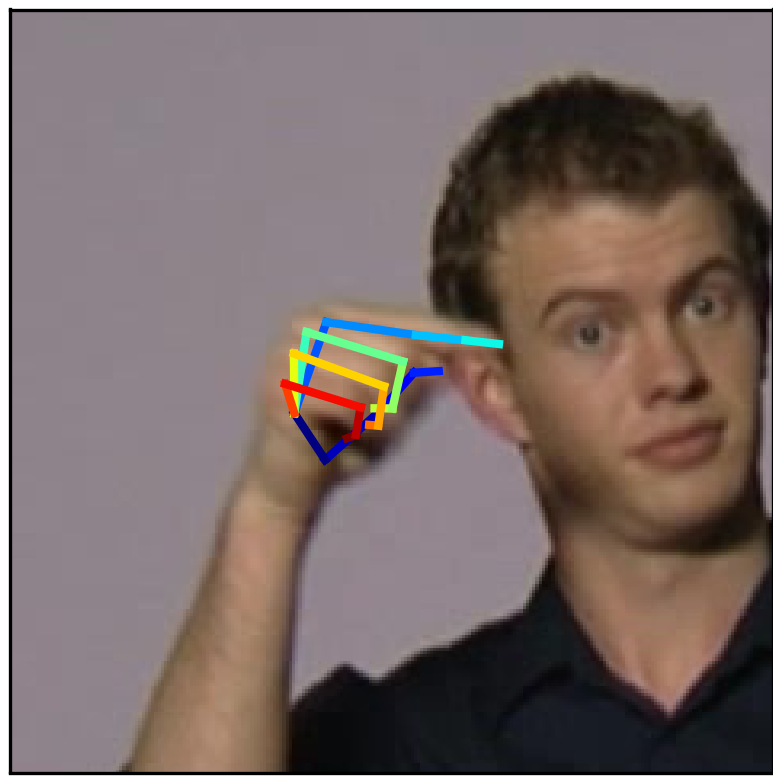} &  
    \includegraphics[trim={10cm 2cm 10cm 2cm},clip,height=0.130\linewidth]{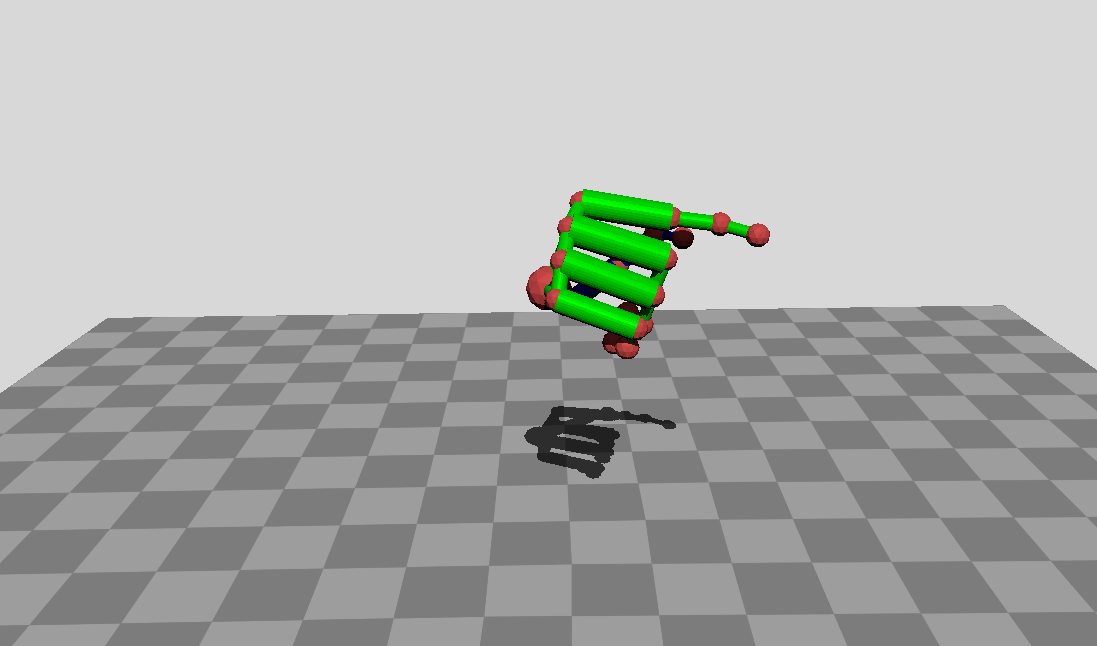} &
    \includegraphics[trim={1cm 1cm 1cm 1cm},clip,height=0.130\linewidth]{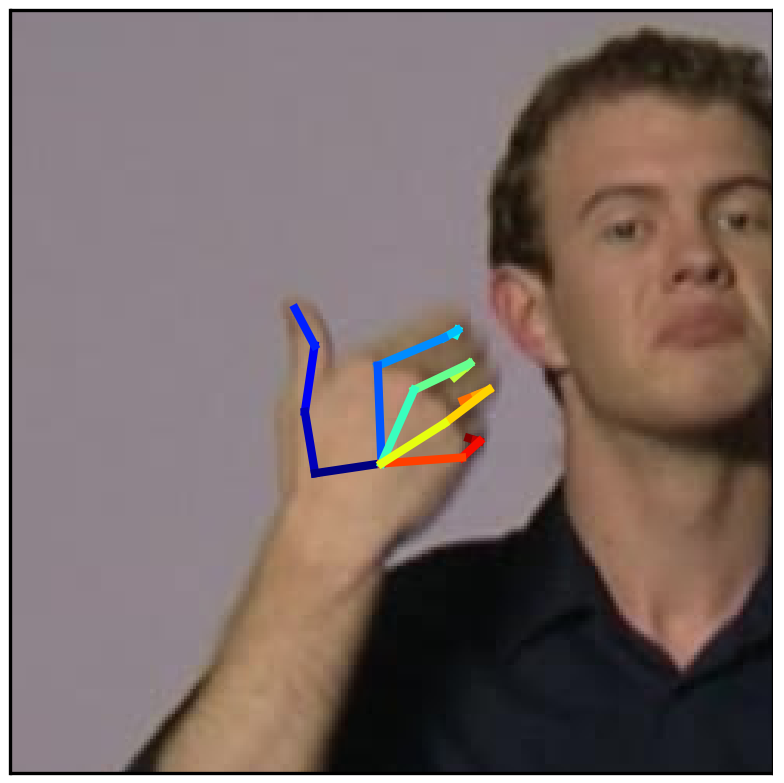}   &
    \includegraphics[trim={10cm 2cm 10cm 2cm},clip,height=0.130\linewidth]{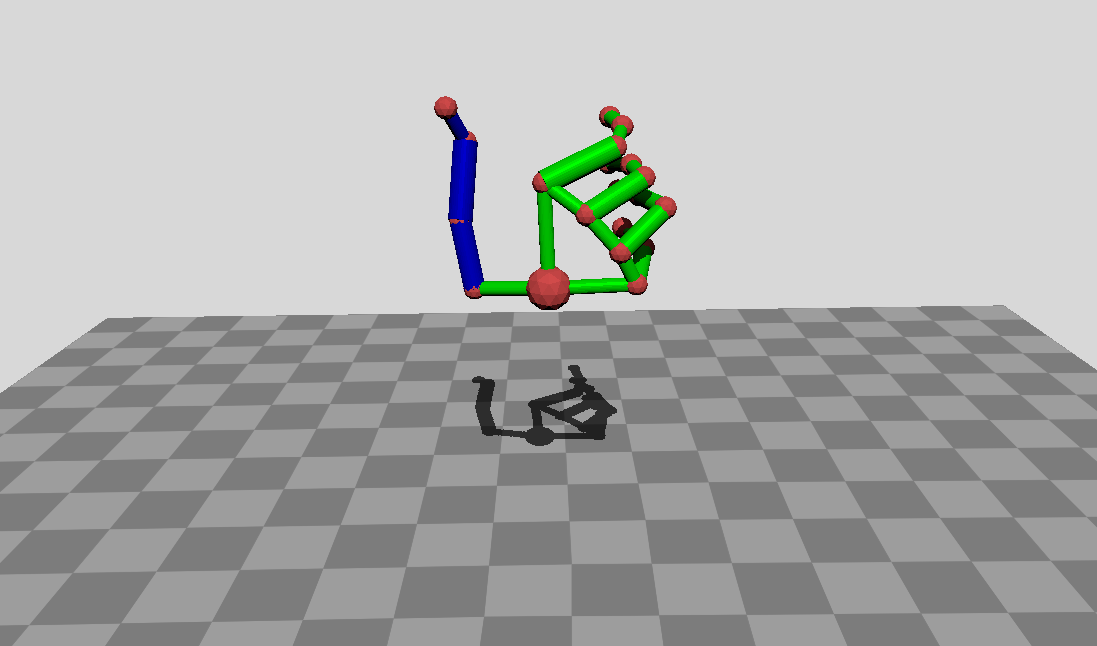} &
    \includegraphics[trim={.85cm 1cm .85cm 1cm},clip,height=0.130\linewidth]{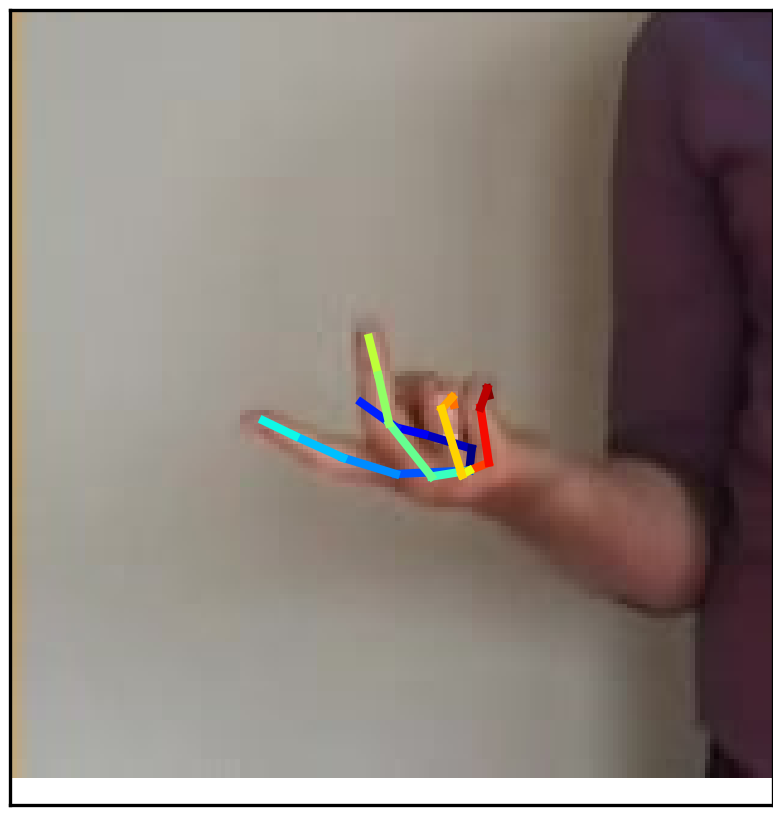}   &
    \includegraphics[trim={7.5cm 1.60cm 11cm 1cm},clip,height=0.130\linewidth]{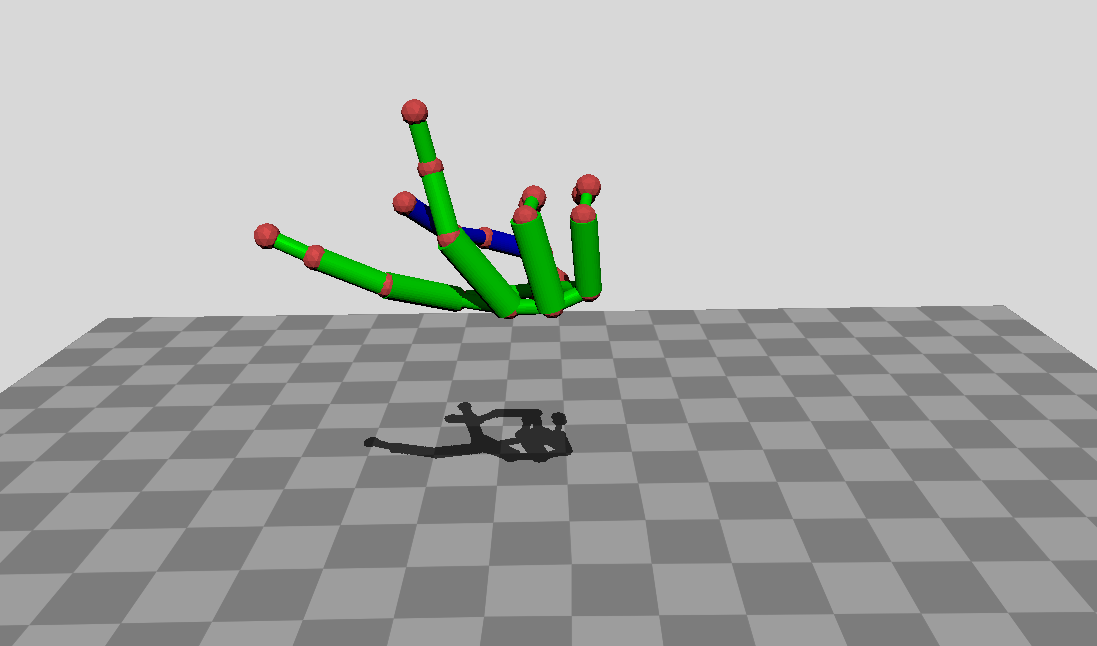}
    \vspace{-1mm} \vspace{-2mm} \\
  \end{tabular}
}
\caption{Qualitative Results. The proposed approach can handle severe occlusions, complex hand articulations, and unconstrained images taken from the wild.\vspace{-5mm}}
\label{fig:qualitative_results}
\end{figure}

Some qualitative results for 3D hand pose estimation for in-the-wild images can be seen in Fig.~\ref{fig:qualitative_results}. 

\vspace{-2mm}
\section{Conclusion}
\vspace{-2mm}
We have presented a method for 3D hand pose estimation from a single RGB image. We demonstrated that the absolute 3D hand pose can be reconstructed efficiently from a single image up to a scaling factor. We presented a novel 2.5D pose representation which can be recovered  easily from RGB images since it is invariant to absolute depth and scale ambiguities. It can be represented as 2.5D heatmaps, therefore, allows keypoint localization with sub-pixel accuracy. We also proposed a CNN architecture to learn 2.5D heatmaps in a latent way using a differentiable loss function. Finally, we proposed an approach to reconstruct the 3D hand pose from 2.5D pose representation. The proposed approach demonstrated state-of-the-art results on five challenging datasets with severe occlusions, object interactions and images taken from the wild. 

\clearpage

\appendix
\section*{Appendix}

In this appendix we provide implementation details to reproduce results in the paper (Sec.~\ref{sec:implementation_details}) and also provide additional ablative studies in Sec.~\ref{sec:ablative}. 

\section{Implementation Details}
\label{sec:implementation_details}
\subsection{Holistic 2.5D Regression}
\label{sec:holistic_regression}
We follow~\cite{sun2017compositional} and use a ResNet-50~\cite{he_cvpr2016} model for holistic regression. As in~\cite{sun2017compositional}, we mean normalize the poses before training and use $L_1$ norm as the loss function. The input to the network is a $224 \times 224$ image. We use $\alpha=1$ since the poses are normalized and the range of $\mathcal{L}_{xy}$ and $\mathcal{L}_{\hat{z}}$ is similar. The initial learning rate is set to $0.03$. 
 
\subsection{2.5D Heatmap Regression}
\begin{figure}[!]
   	  \center
	  \includegraphics[trim={0.5cm 11.2cm .5cm 9cm},clip,width=\linewidth]{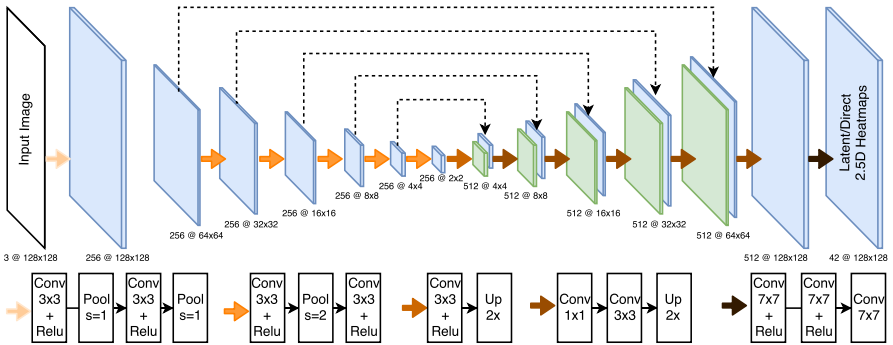}
	  \caption{Backbone network used for 2.5D heatmap regression.\vspace{-3mm}}
      \label{fig:network}
\end{figure}

For 2.5D heatmap regression, we use an Encoder-Decoder architecture with skip connections~\cite{newell2016eccv} and fixed number of channels (256) in each convolutional layer. The detailed network architecture can be seen in Fig.~\ref{fig:network}. The input to our model is a $128 \times 128$ image, which produces full resolution latent/direct 2.5D heatmaps as output. 

\subsubsection{Direct 2.5D Heatmap Regression:}
We use $\sigma=5$ to create the target heatmaps for training. We follow~\cite{wei2016convolutional,newell2016eccv} and use $L_2$ norm as the loss function. The initial learning rate is set to $0.0001$.

\subsubsection{Latent 2.5D Heatmap Regression:}
\begin{figure}[t]
   	  \center
	  \includegraphics[trim={0.0cm 00cm .0cm 0cm},clip,width=\linewidth]{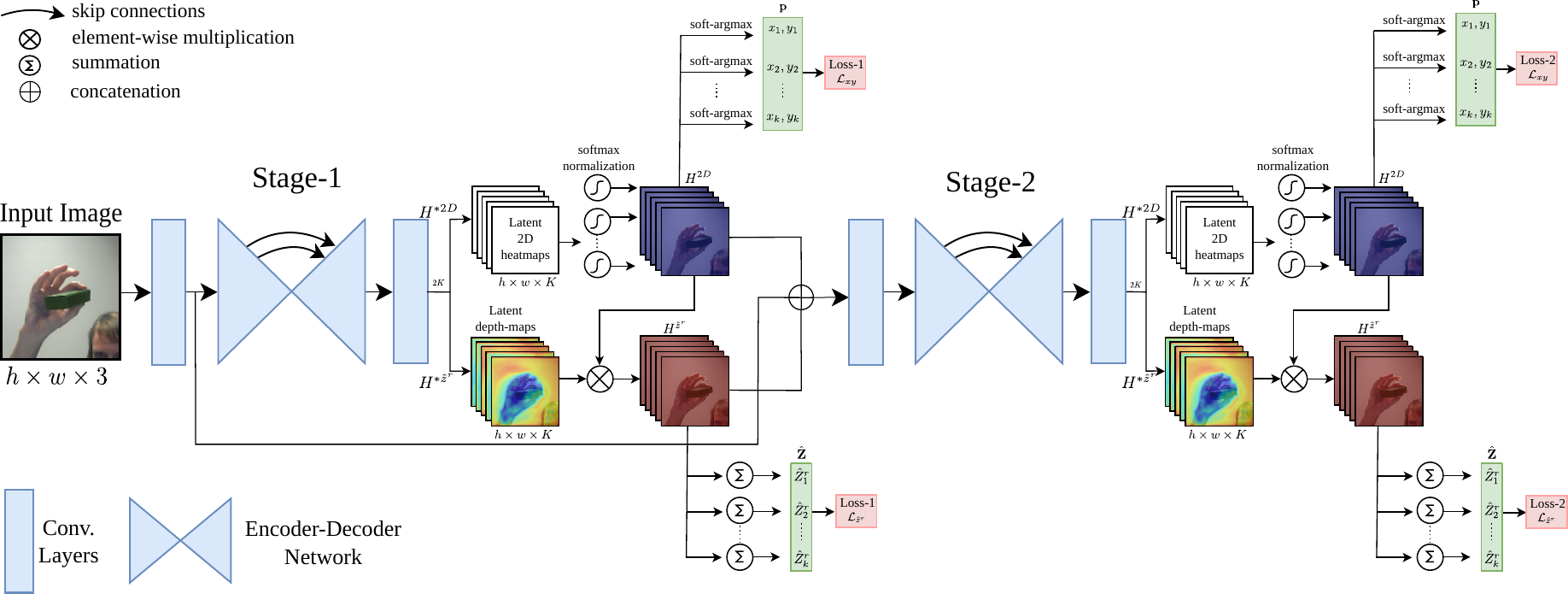}
	  \caption{Overview of the two stage model for latent 2.5D heatmap regression.}
      \label{fig:two_stage_model}
\end{figure}

For latent 2.5D regression, the $\mathcal{L}_{xy}$ is calculated on 2D pixel coordinates and $\mathcal{L}_{\hat{z}^r}$ is computed on the scale normalized root-relative depths. Therefore, a balancing factor is required. We empirically chose $\alpha=20$ such that both losses have a similar magnitude. In our experiments we also tried with $\alpha=1$ and the performance dropped insignificantly by less than $1\%$. We use $L_1$ norm as the loss function with a learning rate of $0.001$. The overview of the two-stage model for 2.5D heatmap regression can be seen in Fig.~\ref{fig:two_stage_model}

\subsection{Common details}
We train the models only for the right hand, and during inference, flip the left hand images before passing them to the network. All models are trained from scratch with a batch size of $32$ for $70$ epochs.
During training, we crop the bounding box such that the hand is ~$70\%$ of the image. We perform data augmentation by rotation $(0,\ang{90})$, translation ($\pm20$ pixel), scale (0.7,1.1), and color transformations. In addition, in order to make the models robust against object occlusions, we follow~\cite{mueller_iccv2017} and randomly add textured objects (ovals and cubes) to the training samples.  
We decay the learning rates for all models by a factor of $10$ after every $30$ epochs, and use SGD with $\mathrm{momentum} = 0.9$. During mixed training, the training images with 2D-only or 3D annotations are sampled with equal probability.

\section{Additional ablative studies}
\label{sec:ablative}
\begin{figure*}[t] 
   \centering
    \begin{subfigure}[t]{0.365\textwidth}
   	  \center
	  \includegraphics[width=\columnwidth]{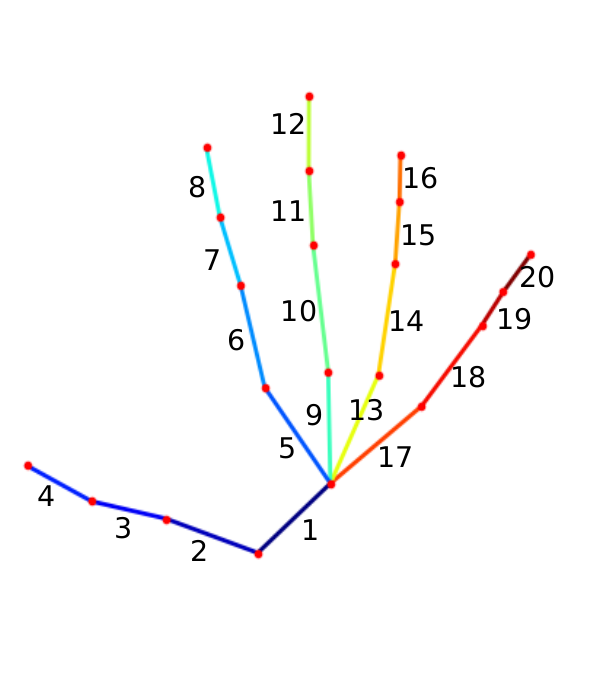}
	  \caption{Skeleton of the hand used in this work with bone ids.}
      \label{fig:skeleton}
      \end{subfigure}
     \hspace{0.3cm}
	\begin{subfigure}[t]{0.45\textwidth}
   	  \center
	  \includegraphics[width=\columnwidth]{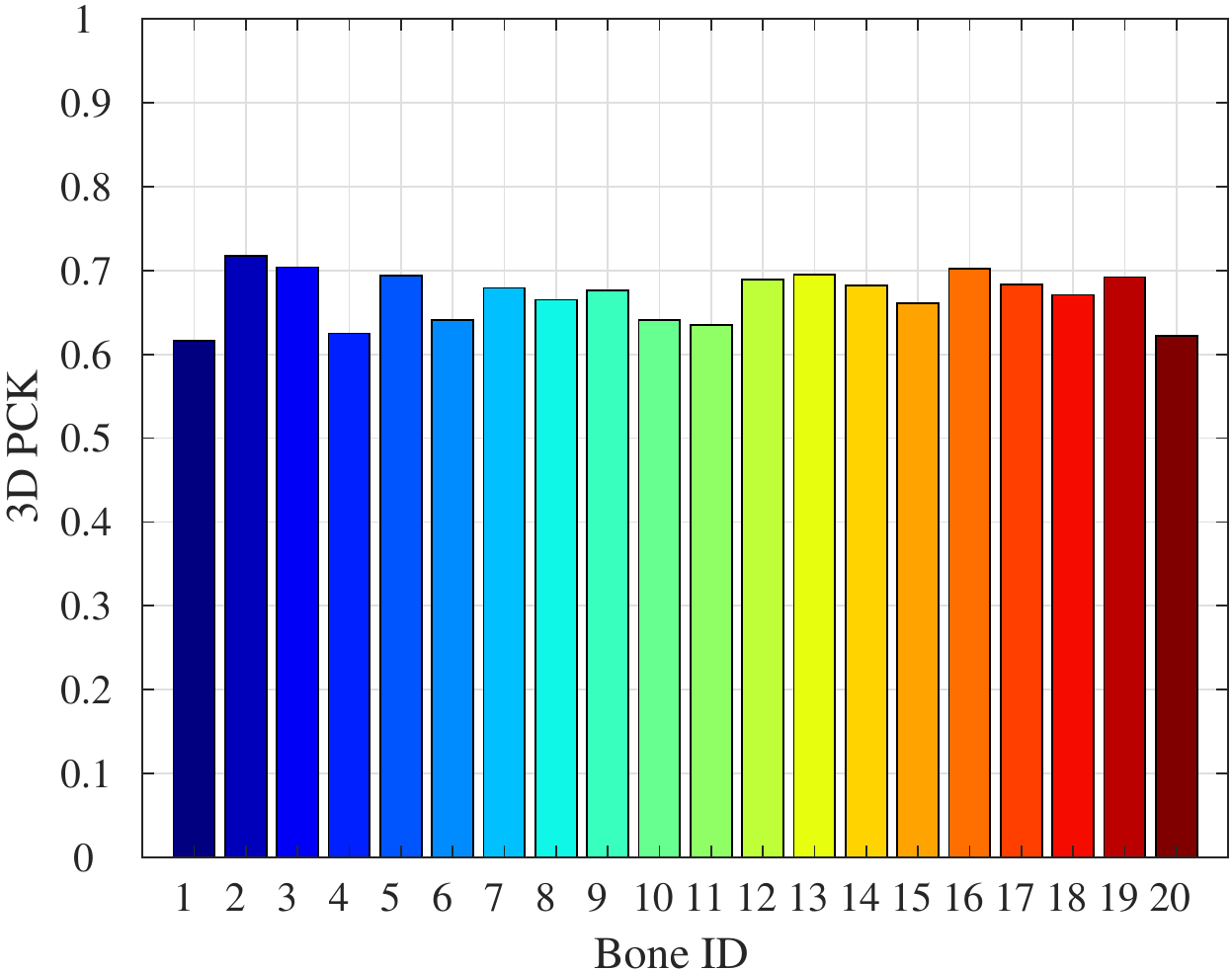}
	  \caption[short]{Impact of the bone used for normalization in eqt.~\ref{eqt:projection3d2d} and for reconstruction of 3D pose from 2.5D (Sec.~\ref{sec:3D_reconstruction}).}
      \label{fig:impact_of_bones}
	\end{subfigure}
  \caption{Additional ablative studies.}
\end{figure*}

The skeleton of the hand used in this work can be seen in Fig.~\ref{fig:skeleton}. We evaluate the impact of pair of keypoints (bones) selected for 3D pose normalization (eqt.~\ref{eqt:normalization} in Fig.~\ref{fig:impact_of_bones}. For this, we trained a separate CNN model while using a specific pair of keypoints for normalization. We can see that the performance remains consistent ($\approx 0.69$) for most of the bones.

\bibliographystyle{splncs}
\bibliography{handbib,pose3d}

\clearpage

\end{document}